\documentclass[times,twocolumn,final]{elsarticle}
\usepackage{medima}
\usepackage{framed}
\usepackage{subfiles}
\usepackage{amssymb}
\usepackage{amsmath}
\usepackage{latexsym}
\usepackage{bm}
\usepackage{graphicx}
\usepackage{subcaption}
\usepackage{caption}
\usepackage{booktabs}
\usepackage{tabularx}
\usepackage{multirow}
\usepackage{arydshln}
\usepackage{makecell}
\usepackage{array}
\usepackage{algorithm}
\usepackage{algpseudocode}
\usepackage{xcolor}
\usepackage{ulem}
\usepackage{url}
\usepackage{geometry}

\definecolor{darkblue}{rgb}{0.0, 0.0, 0.55}
\usepackage[colorlinks=true,linkcolor=darkblue,urlcolor=darkblue,citecolor=darkblue]{hyperref}

\journal{Medical Image Analysis}

\begin{document}
	
	\verso{X. Zhang \textit{et~al.}}
	
	\begin{frontmatter}
		
		\title{Non-Contrast CT Esophageal Varices Grading through Clinical Prior-Enhanced Multi-Organ Analysis}%
		\author[label1,label2,label4,label5]{Xiaoming Zhang\fnref{equal}}
		\author[label1]{Chunli Li\fnref{equal}}
		\fntext[equal]{These authors contributed equally to this work.}
		\author[label3]{Jiacheng Hao}
		\author[label2,label5]{Yuan Gao}
		\author[label2,label5]{Danyang Tu}
		\author[label1]{Jianyi Qiao}
		\author[label1]{Xiaoli Yin}
		\author[label2]{Le Lu}
		\author[label2]{Ling Zhang}
		\author[label2,label5]{Ke Yan}
		\author[label1]{Yang Hou}
		\author[label1]{Yu Shi}
		\cortext[*]{Corresponding author e-mail: 20072265@cmu.edu.cn (Yu Shi); houy2@sj-hospital.org (Yang Hou)}
		
		\address[label1]{Department of Radiology, Shengjing Hospital of China Medical University, 110004, Shenyang, China}
		\address[label2]{DAMO Academy, Alibaba Group}
		\address[label3]{School of Biomedical Engineering, Tsinghua University, 100084, Beijing, China}
		\address[label4]{Faculty of Science and Engineering, Sorbonne University, 75005, Paris, France}
		\address[label5]{Hupan Lab, 310023, Hangzhou, China}		
		\received{2 February 2025}
		\finalform{20 December 2025}
		\accepted{21 December 2025}
		%\availableonline{**}
		%\communicated{**}

		\begin{abstract}
			Esophageal varices (EV) represent a critical complication of portal hypertension, affecting approximately 60\% of cirrhosis patients with a significant bleeding risk of $\sim$30\%. While traditionally diagnosed through invasive endoscopy, non-contrast computed tomography (NCCT) presents a potential non-invasive alternative that has yet to be fully utilized in clinical practice. We present Multi-Organ-COhesion Network++ (MOON++), a novel multimodal framework that enhances EV assessment through comprehensive analysis of NCCT scans. Inspired by clinical evidence correlating organ volumetric relationships with liver disease severity, MOON++ synthesizes imaging characteristics of the esophagus, liver, and spleen through multimodal learning. We evaluated our approach using 1,631 patients, those with endoscopically confirmed EV were classified into four severity grades. Validation in 239 patient cases and independent testing in 289 cases demonstrate superior performance compared to conventional single organ methods, achieving an AUC of 0.894 versus 0.803 for the severe grade EV classification (G3 versus $<$G3) and 0.921 versus 0.793 for the differentiation of moderate to severe grades ($\geq$G2 versus $<$G2). We conducted a reader study involving experienced radiologists to further validate the performance of MOON++. To our knowledge, MOON++ represents the first comprehensive multi-organ NCCT analysis framework incorporating clinical knowledge priors for EV assessment, potentially offering a promising non-invasive diagnostic alternative. Code is available at \href{https://github.com/StevenHaojc/MOON}{https://github.com/StevenHaojc/MOON}.
			%%%%
		\end{abstract}
		
		
	\end{frontmatter}
	
	\section{Introduction}
	%\renewcommand{\thefootnote}{\textasteriskcentered}
	%\footnotetext{Xiaoming Zhang and Chunli Li contribute equally to this work. Corresponding author Yushi, Email address 18940259980@163.com}
	%\cortext[cor1]{Corresponding author.}
	\begin{figure}[ht!]
		\centering
		\centering
		\includegraphics[width=1\linewidth]{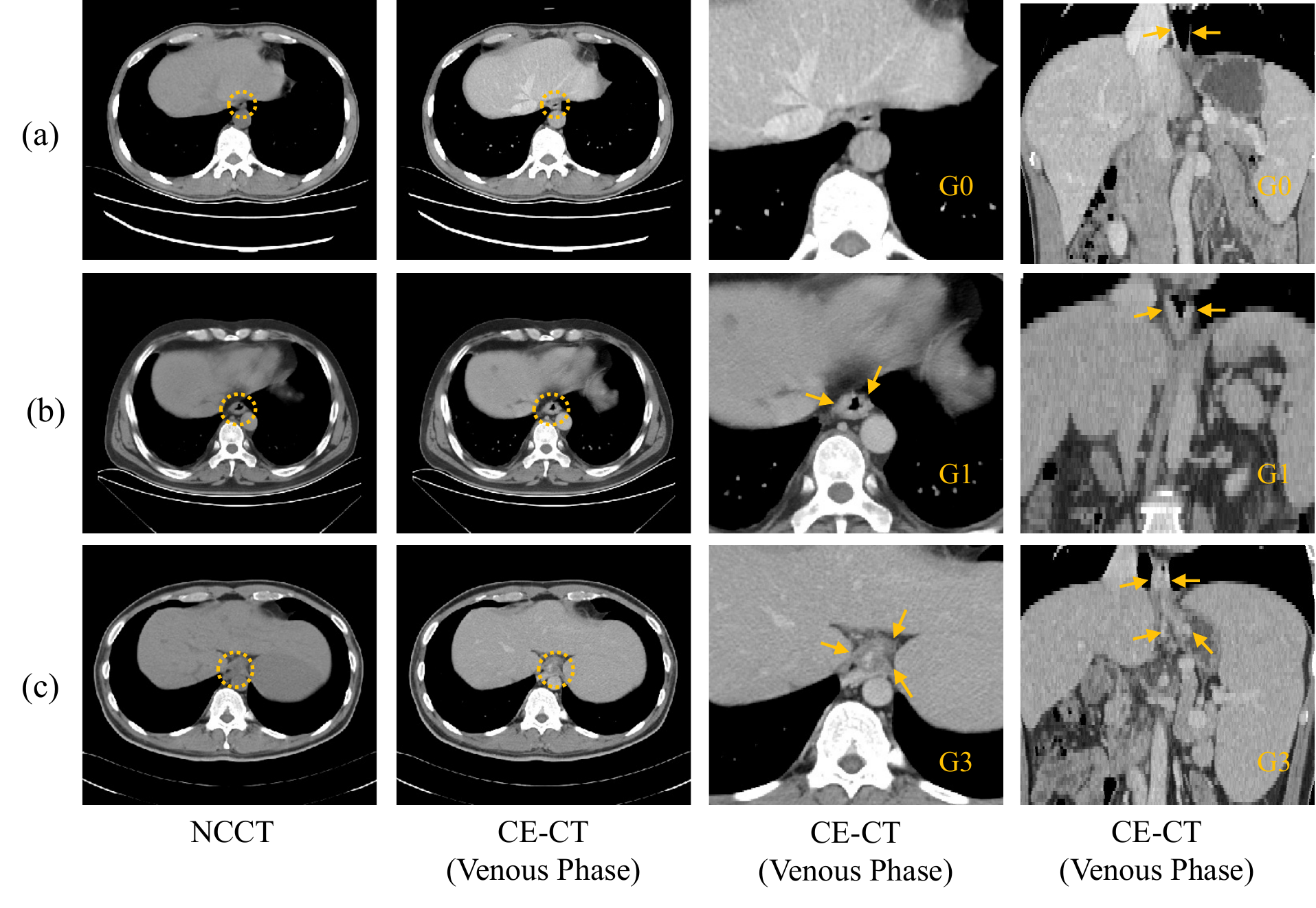}
		\caption{Demonstration of esophageal varices grading on non-contrast and contrast-enhanced CT scans with registration~\citep{tian2023same}. (a) Grade 0: No EV present, showing normal esophageal thickness without dilation. (b) Grade 1: Mild EV, not visible on non-contrast CT but apparent on venous phase CT post-contrast, displaying mild thickening and mildly tortuous vascular shadows. (c) Grade 3: Severe EV with dilated variceal clusters compressing the stomach fundus, showing esophageal narrowing and heterogeneous enhancement.}
		\label{fig:EV}
	\end{figure}
	
	Esophageal varices (EV) are significant complications that often arise from chronic liver diseases, such as liver cirrhosis. EVs are present in about 50\% of cirrhosis patients and up to 85\% of those with Child-Pugh class C cirrhosis. Acute variceal bleeding occurs in 5\%-15\% of individuals with varices annually. Despite advances in diagnostics and therapy, the six-week mortality rate from variceal bleeding can reach 20\%~\citep{kapoor2015endoscopic}. Portal hypertension is a key factor that drives blood into smaller collaterals at the junction of the lower esophagus and the stomach fundus~\citep{garcia2017portal}. This condition can cause life-threatening hemorrhages and shock, associated with high rates of mortality and morbidity~\citep{luo2023clinical}. Detecting EV early is challenging and typically requires invasive endoscopic procedures for confirmation. Currently, non-invasive predictive models are not widely available in clinical practice. As the extent of venous dilation is correlated with the risk of bleeding, a precise assessment of the severity of EV is crucial. Early intervention and prompt diagnosis are essential for effective and timely management of these EV positive patients~\citep{kapoor2015endoscopic,gralnek2022endoscopic}.

	The current diagnostic landscape for EV includes two primary approaches. Endoscopy, or esophagogastroduodenoscopy, remains the clinical gold standard. It allows for direct visualization of variceal characteristics and assessment of bleeding risk indicators~\citep{kapoor2015endoscopic,dong2019machine}. However, this procedure is invasive, posing significant patient burdens such as infection and bleeding risks, as well as increased healthcare costs. Alternatively, contrast-enhanced CT (CE-CT) has emerged as a less invasive option, offering comprehensive visualization of varices and associated vessels~\citep{yan2022novel}. Despite its advantages, methods leveraging the radiomic characteristics of CE-CT~\citep{yan2022novel} face limitations in generalizability. Furthermore, CE-CT involves higher radiation exposure and the use of iodinated contrast agents, which can lead to adverse reactions.
	
\begin{figure}[ht!]
		\centering
		\centering
		\includegraphics[width=1\linewidth]{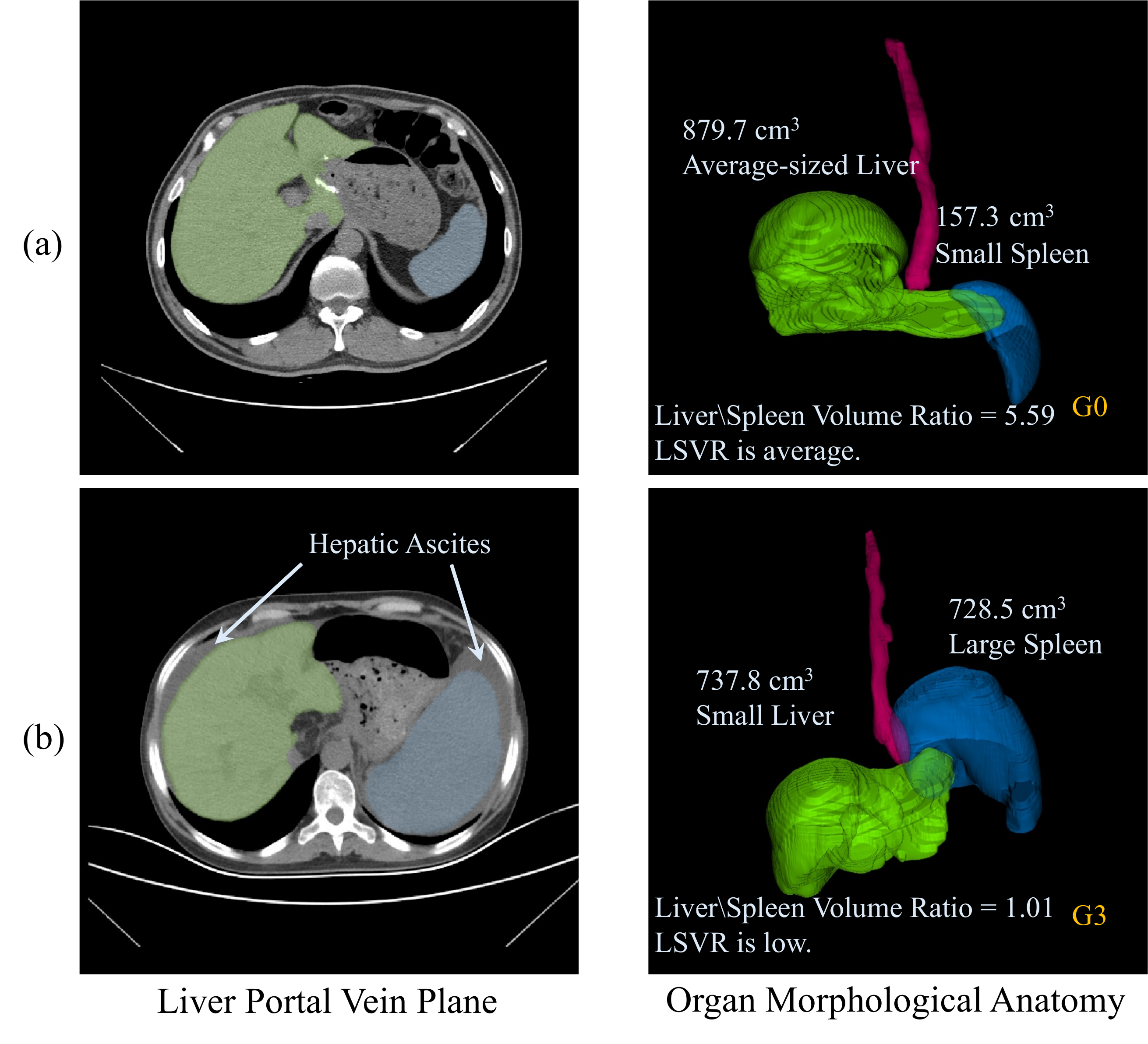}
		
		\caption{Comparison of organ anatomy in EV: (a) Non-EV subject displaying normal liver and spleen volume. (b) Subject with severe EV (Grade 3), commonly associated with decompensated cirrhosis. In the spleen, enlargement often occurs due to portal vein hypertension. In the liver, several changes can be observed, e.g., alterations in liver density resulting from abnormal perfusion and the presence of hepatic ascites surrounding the abdominal cavity. Clinically, LSVR is a more effective factor in evaluating liver fibrosis.}
		\label{fig:Anato}
	\end{figure}

	Compared with CE-CT, non-contrast CT (NCCT) offers a quick and low-radiation alternative and valuable for the clinical evaluation of EV. As shown in Figure~\ref{fig:EV}, mild EV are not visible on non-contrast CT. For patients with liver cirrhosis, the esophagus imaging can appear similar between those without EV and those with mild EV, since only a contrast agent can reveal the deformed vessels. Liver characteristics such as fibrosis and volume alterations can indicate portal venous pressure, which is crucial for assessing the risk and severity of EV~\citep{shi2021three}. As shown in Figure~\ref{fig:Anato}, changes in spleen volume caused by portal hypertension, such as enlargement, indirectly reflect this pressure and could suggest a risk of EV bleeding. Therefore, a combined analysis of the liver, spleen, and esophagus can improve understanding and grading of the risks of EV and strengthen clinical decision-making. However, several challenges persist in the effective and accurate assessment of EV: difficulties in differentiation due to low CT contrast resolution, limited visibility of smaller varices, and inconsistent distribution of varices throughout the esophagus.
	
	Previous studies on evaluating EV have focused on CE-CT imaging of vascular characteristics of the liver, spleen, esophagus, and portal venous system~\citep{luo2023clinical,wan2022quantitative,yan2022novel}. However, few grading methods fully consider the close relationships of the esophagus with adjacent anatomical structures. To optimize EV evaluation, we have introduced the multi-organi-cohesion network (MOON)~\citep{li2024improved} framework based on chest NCCT. Our approach improves the accuracy of the EV diagnosis and surpasses previous radiomic methods that rely on CE-CT imaging.
	
	MOON adopts a holistic strategy, simultaneously analyzing the regions of interest (ROIs) from NCCT scans of these related organs. It employs a multi-organ configuration that incorporates the Organ Representation Interaction (ORI), allowing features from the liver and spleen to inform the analysis of esophageal characteristics within the feature extraction. To achieve uniformity in decision making across various organ domains, it incorporates ordinal regression loss~\citep{cheng2008neural} and canonical correlation analysis (CCA)~\citep{andrew2013deep} loss into the training process. This synergistic use of loss functions fosters a cohesive feature integration, thus reducing inconsistencies in the decision-making process and increasing overall diagnostic performance.

	Based on MOON's development, we propose MOON++, which incorporates clinical priors derived from key organ volumes, established indicators of liver cirrhosis severity, to enhance EV staging accuracy. We investigate optimal methods for integrating these clinical priors into image backbones, comparing medical multi-modal large language models (MLLM) approaches~\citep{bai2024m3d} with more straightforward techniques. We validated our approach on an extended dataset of 1,474 patients, comprising both EV-positive and EV-negative cases. The endoscopically confirmed EV cases were further categorized into four severity grades. MOON++ demonstrates exceptional diagnostic capability, achieving AUC values of \textbf{0.894} for severe EV classification (compared to 0.736 with CE-CT~\citep{yan2022novel}) and \textbf{0.921} for moderate-to-severe case detection (versus 0.802 with CE-CT~\citep{wan2022quantitative}).
	
	Although NCCT is not typically considered the recommended modality for staging EV due to several aforementioned challenges, to our knowledge, MOON~\citep{li2024improved} is the first work to incorporate a synchronized multi-organ analysis on NCCT for EV assessment, offering a more accessible alternative for patients compared to traditional, invasive modalities, e.g., CE-CT and endoscopy. This work expands on our conference paper version at MICCAI 2024~\citep{li2024improved}, building on our previous work, the novel contributions of this paper are as follows:
	
	\begin{itemize}
		\item We expanded the dataset to include 1,631 samples by adding subjects in various EV grades and external validation, including 218 subjects without EV (i.e. G0), thus increasing the diversity of the sample population.
		\item We streamlined the multi-organ image backbone by utilizing more effective switching interaction strategies and enhanced the CCA loss with deep CCA (DCCA).
		\item We validated our approach through a reader study involving two experienced radiologists, providing crucial clinical validation.   
		\item We leverage key organ volume relationships as clinical priors to enhance multi-organ cohesion for improving EV grading. We also examined whether medical multi-modal large language models are necessary for information encoding or if simpler embedding methods could be sufficient.
		\item Extensive experiments were conducted on the expanded dataset, including more comparative studies. 
	\end{itemize}
	
	The remainder of the article is organized as follows: Section 2 reviews related work in EV assessment; Section 3 \& 4 presents the MOON++ pipeline, experimental setup, evaluation results and visualization analysis; and Section 5 discusses and concludes the study.
	\section{Related Work}
	Recent studies have demonstrated significant progress in applying deep learning techniques to evaluate EV. The development has evolved from traditional clinical assessment methods to sophisticated machine learning approaches that integrate multiple data sources and modalities.
	
	\subsection{Clinical Assessments}
	In the clinical evaluation of EV, both the European Society of Gastrointestinal Endoscopy (ESGE)~\citep{gralnek2022endoscopic} and the Japan Society of Portal Hypertension (JSPH)~\citep{kawano2008short} provide standardized guidelines. Although ESGE uses Child-Pugh score (CPS) and the Model for End Stage Liver Disease (MELD) score for risk stratification, and JSPH categorizes EV based on varix form, these traditional methods are invasive and potentially risky for certain patients. The liver-to-spleen volume ratio (LSVR) is a valuable objective metric that negatively correlates with the CPS and the MELD score~\citep{kwon2021liver}, both of which assess liver function and severity of cirrhosis. Inspired by its operator-independent nature and easy accessibility, we incorporate LSVR and organs volumes as the text prompt in our multi-modal integration framework for automated EV assessment.
	
	Recent advances in medical image analysis have demonstrated the potential of deep learning in analysis for various gastrointestinal and hepatic conditions, e.g., splenomegaly segmentation on multi-modal MR images~\citep{huo2018splenomegaly}, feature decoupling techniques for fatty liver detection using ultrasound~\citep{huang2024robustly}, and interpretable multi-view learning for liver fibrosis staging on MR images~\citep{gao2023reliable}. 
	\subsection{Radiomics-based Methods}
	Dong et al.~\citep{dong2019machine} developed the EVendo score using random forest algorithms on clinical parameters, achieving AUC scores of 0.84 and 0.82 in training and validation datasets, respectively. Huang et al.~\citep{huang2023development} created a light gradient boosting model using liver stiffness, platelet count, and total bilirubin, which demonstrated superior performance compared to traditional Baveno VI criteria~\citep{de2015expanding} in avoiding unnecessary endoscopies. The evolution of CT-based EV assessment methods has been well documented in earlier work. Yang et al.~\citep{yang2019predicting} established the basis for using CT radiomic signatures to predict variceal esophageal bleeding, particularly in cases related to hepatitis B-induced cirrhosis. Their integrated model achieved an AUC of 0.83, demonstrating the potential to combine radiomics with clinical characteristics. Recent systematic reviews have highlighted both the promise and limitations of these approaches, e.g., Malik et al.~\citep{malik2024systematic} examined various machine learning models, noting impressive precision in some cases. However, significant challenges remain, including large-scale validation, standardization methodologies, and potential biases. In particular, compared to endoscopy, CT imaging is viewed as a more standardized approach, as it does not heavily depend on the expertise of the individual operator.
	
	\subsection{Deep Learning-based Methods}
	Compared to radiomics approaches, deep learning-based methods for assessing EV remain underexplored. Early deep learning applications focused on liver diseases, particularly fibrosis staging using CE-CT, with Yasaka et al.\cite{yasaka2018deep} pioneering CNN-based approaches and Yin et al.\cite{yin2021liver} achieving improved interpretability. For EV bleeding risk assessment, significant progress has been made in organ relationship modeling using CE-CT, the Liver-Spleen model~\citep{gao2023imaging} achieved AUCs of 0.789 on CE-CT, while~\citep{luo2023clinical} further enhanced performance by integrating radiomics features with clinical variables.
	
	Recent advances in multi-modal fusion have demonstrated effectiveness across various esophageal conditions. Zhang et al.~\citep{zhang2024multi} proposed a dual-layer cross-attention model for esophageal fistula prognosis, addressing the heterogeneity between clinical data and CT images. Wu et al.~\citep{wu2024mmfusion} introduced a graph-based multi-modal diffusion model for lymph node metastasis diagnosis, tackling information redundancy challenges. In endoscopic analysis, Hou et al.~\citep{hou2021early} developed a hierarchical aggregation approach for early neoplasia identification, complemented by advances in Transformer~\citep{kusters2025will} and robust feature learning techniques~\citep{jaspers2024robustness} for subtle anatomical change detection.
	
	For NCCT specifically, architectural innovations have emerged, including position-sensitive attention mechanisms~\citep{yao2022effective} and comprehensive approaches for tumor volume segmentation and diagnosis~\citep{jin2021deeptarget,hao2025plus}. However, the potential of NCCT for EV assessment remains underexplored, particularly in approaches integrating multi-organ analysis with clinical knowledge. While current methods predominantly rely on CE-CT or endoscopy, there exists a significant opportunity to leverage NCCT's accessibility and safety advantages through advanced deep learning techniques.
	
	\section{Method}

	\begin{figure*}[ht!]
		\centering
		\begin{subfigure}{1\linewidth} 
			\centering
			\includegraphics[width=0.9\linewidth]{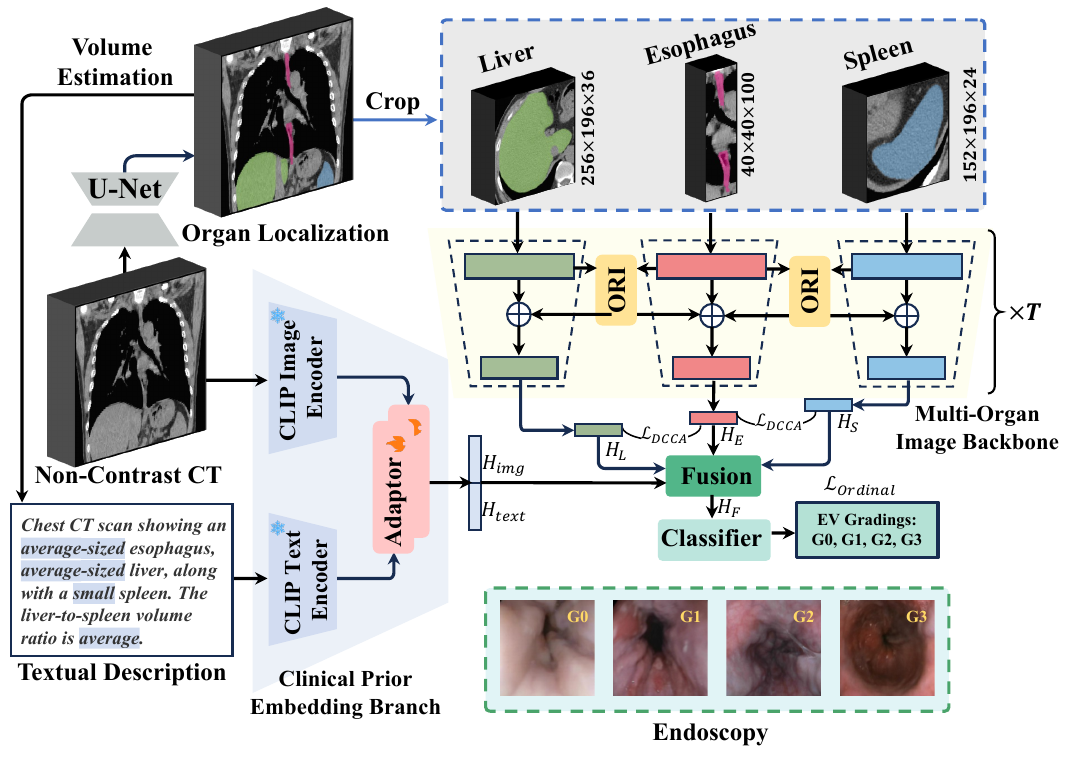} 
			\label{fig:pipe_a}
		\end{subfigure}
		
		\vspace{0.2em} 
		
		\begin{subfigure}{1\linewidth}
			\centering
			\includegraphics[width=0.9\linewidth]{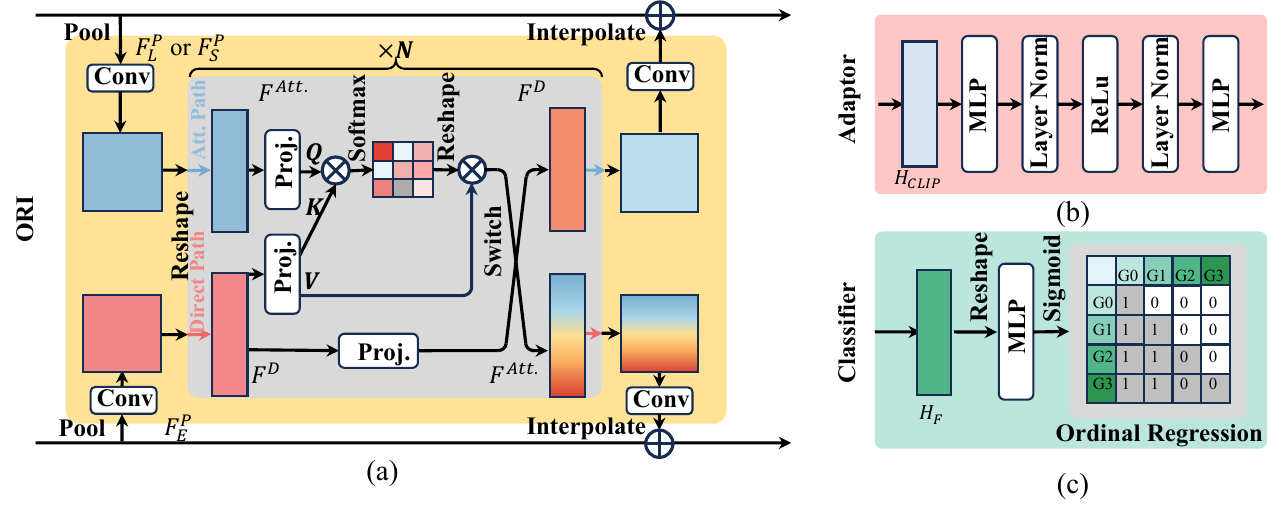}  
			\label{fig:pipe_b}
		\end{subfigure}
		\caption{Overview of the Multi-Organ Cohesion Network++. (a) Organ representation interaction. (b) Adaptor. (c) Classifier.}
		\label{fig:pipe}
	\end{figure*}
	Deep learning, particularly CNNs ~\citep{he2016deep, lecun1989handwritten, lecun1989backpropagation}, has significantly advanced the detection of pathologies and diseases that exhibit subtle imaging differences in Hounsfield units (HU) on NCCT images, often imperceptible to radiologists~\citep{wasserthal2023totalsegmentator}. MOON++ features four key components: (1) Assessing EV requires a global receptive field~\citep{yao2022effective}. We use UniFormer~\citep{li2023uniformer}, which combines convolution with self-attention, to effectively capture esophageal features for accurate EV evaluation. (2) EV assessment requires integrating feature representations from key organs, i.e., liver and spleen. The Organ Representation Interaction (ORI) module achieves this by combining features from these organs with esophageal features, enhancing EV detection. (3) To enhance precise assessment of EV severity grades across multiple organ branches, we employ a hybrid loss function combining ordinal regression with DCCA, as detailed in Algorithm~\ref{dcca}. DCCA preserves ordinal relationships in EV severity levels while optimizing inter-organ feature correlations. (4) To leverage clinical priors, we incorporate an embedding branch that processes organ measurements (volumes and LSVR), enriching the model with established clinical indicators for more precise EV staging.
	
	\subsection{Multi-Organ-cOhesion-Network++ (MOON++)}
	
	\noindent\textbf{Pipeline of MOON++.}  MOON++, as illustrated in Figure~\ref{fig:pipe}, combines multi-organ image backbones with a clinical prior embedding branch. The framework begins with a pre-trained nnUNet~\citep{isensee2021nnu} for organ segmentation, followed by UniFormer~\citep{li2023uniformer} backbones that facilitate both local and global feature extraction. The ORI module processes organ-specific characteristics essential for EV assessment. The clinical prior embedding branch incorporates recent medical MLLM M3D~\citep{bai2024m3d}, which employs a CLIP-like training strategy~\citep{radford2021learning}. M3D encoders process organ volumes and LSVR information, and features from all branches are ultimately integrated for four-stage EV classification.
	
	\noindent\textbf{Organ Representation Interaction.}
	To address the diagnosis of EV in portal hypertension, comprehensive CT analysis of the liver and spleen is essential due to their subtle morphological and textural changes. As illustrated in Figure~\ref{fig:pipe}\textcolor{red}{(a)}, the ORI module refines and integrates the imaging features of the liver and spleen while directing attention to critical areas. Given the input features of the esophagus $\textbf{F}_{E}\in\mathbb{R}^{H_{e}\times W_{e}\times D_{e} \times C}$, liver $\textbf{F}_{L}\in\mathbb{R}^{H_{l}\times W_{l}\times D_{l} \times C}$, and spleen $\textbf{F}_{S}\in\mathbb{R}^{H_{s}\times W_{s}\times D_{s} \times C}$, the module first pools these features to uniform dimensions: $\textbf{F}_{E}^{P}, \textbf{F}_{L}^{P}, \textbf{F}_{S}^{P} \in \mathbb{R}^{H_{n}\times W_{n}\times D_{n} \times C}$. The pooled features undergo two parallel paths: an attention path and a direct path. In the attention path, the input is reshaped to attention-ready form $F^{\text{Att.}} \in \mathbb{R}^{H_{n}W_{n}\times D_{n}\times C}$ and processed through matrices of query ($Q$), key ($K$) and value ($V$). The scaled dot-product attention is applied as $\text{Attention}(Q, K, V) = \text{SoftMax}\left(\frac{QK^T}{\sqrt{d_k}}\right)V$, with the outputs projected into an enhanced feature space via $F^{\text{Att.}} = TW_P$, where $W_P \in \mathbb{R}^{C\times C}$. The direct path applies linear projection without attention. These paths alternate in a switching manner across $N$ iterations before the final convolution and interpolation steps. As demonstrated in Table~\ref{tab:interaction}, this design reduces computational complexity compared to direct self-attention while improving feature representation.
	
	\noindent\textbf{Clinical Prior Embedding.}
	To enhance the precision of EV classification, we leverage additional clinical information, specifically esophagus, liver and spleen volumes, along with LSVR, as these metrics reflect the degree of liver fibrosis, a condition closely associated with the severity of EV. The organ volumes $V_O$ are computed from segmentation masks and CT voxel spacing parameters using:
	
	\begin{equation}
		V_O = N_O \times (p_x \times p_y \times p_z), \quad O \in {L,S,E},
		\label{volume}
	\end{equation}where $N$ denotes the total number of voxels aggregated across allconnected components in the organ segmentation mask, and $p_x$, $p_y$, $p_z$ represent the physical voxel resolution (in mm) along their respective axes. This volumetric computation is implemented as a computationally efficient post-processing step on the segmentation outputs, requiring no additional model inference overhead. We implement this through medical CLIP encoders based on M3D~\citep{bai2024m3d} to integrate both the complete CT data and the organ text descriptions. Using $V_O$, descriptive text prompts that characterize the organ volumes and their relationships are generated. The encoded features undergo refinement through two separate straightforward adaptors, as illustrated in Figure~\ref{fig:pipe}\textcolor{red}{(b)}, mainly aimed at reducing the dimensionality of the features. The resulting refined features, $\textbf{H}_{img}$ and $\textbf{H}_{text}$, are then concatenated for subsequent fusion. This multimodal approach demonstrates significant improvements in classification performance, yielding approximately 9.5\% increase in overall multi-class accuracy on the validation dataset and 2.7\% improvement on the test dataset.
	
	\noindent\subsection{Training Paradigm of MOON++}
	MOON++ is proposed and designed to tackle the intricacies of EV diagnosis by integrating multi-organ information. Within this scope, we implement an ordinal regression loss, symbolized as $\mathcal{L}_{\text{Ordinal}}$, as shown in Figure~\ref{fig:pipe}\textcolor{red}{(c)}, which is applied to the fusion logits $\mathbf{H}_F$. These logits consolidate data from the esophagus, liver, and spleen branches, according to~\citep{cheng2008neural}. The ordinal regression loss is crucial for measuring the model's ability to accurately classify the different stages of EV, as it verifies the alignment between $\mathbf{H}_F$ and the ground-truth labels $\mathbf{Y}$. Furthermore, MOON++ incorporates the DCCA to enhance CCA from multi-view learning. This approach maximizes the correlation between two feature sets to improve cross-modal retrieval. Compared with CCA~\citep{andrew2013deep,li2024improved}, DCCA directly computes correlations in the projected space without relying on eigenvalue decomposition, offering more flexible representation learning. Specifically, it targets the alignment of logits from the esophageal branch ($\mathbf{H}_E$) with those from the liver ($\mathbf{H}_L$) and spleen ($\mathbf{H}_S$) branches. By serving as a regularization mechanism, $\mathcal{L}_{\text{DCCA}}$ facilitates the convergence of characteristics derived from multiple organs, thus improving the model's ability to interpret interrelated organ information.
	
	The ordinal regression loss $\mathcal{L}_{\text{Ordinal}}$ is formulated as:
	\begin{equation}
		\mathcal{L}_{\text{Ordinal}} = \|\mathbf{H}_F - \mathbf{Y}_{\text{ord}}\|_2^2,
	\end{equation}
	where $\mathbf{H}_F$ represents the model predictions and $\mathbf{Y}_{\text{ord}}$ is the ordinal encoding of ground truth labels. For a 4-class regression (G0-G3), the ordinal encoding transforms each label into a binary vector as shown in Figure~\ref{fig:pipe}\textcolor{red}{(c)}.

	The total loss $\mathcal{L}_{\text{Overall}}$ for a given batch is a composite of the $\mathcal{L}_{\text{Ordinal}}$ and $\mathcal{L}_{\text{DCCA}}$, weighted appropriately, that not only accounts for accurate staging of EV but also reinforces the interconnectedness of the multi-organ representations:
	
	\begin{equation}
		\mathcal{L}_{\text{Overall}} = \lambda \mathcal{L}_{\text{Ordinal}} + (1-\lambda) \sum_{O \in \{L,S\}} \mathcal{L}_{\text{DCCA}}(\mathbf{H}_E, \mathbf{H}_O).
		\label{lcca}
	\end{equation}
	\begin{algorithm}[ht!]
		\caption{Deep Canonical Correlation Analysis Loss}
		\begin{algorithmic}
			\Statex \textbf{Input}: Two relevant feature vectors $H_{1}, H_{2}\in\mathbb{R}^{n,d}$, projection networks $f_1, f_2$, $\epsilon=10^{-12}$.
			\Statex \textbf{Output}: Deep CCA loss $\mathcal{L}_{DCCA}(H_{1}, H_{2})$.
			
			\State $H_{1}, H_{2} \gets f_1(H_{1})$,  $f_2(H_{2})$ \Comment{View non-linear projection}

			\State $H_{1}, H_{2} \gets \frac{H_{1} - \text{mean}(H_{1})}{\text{std}(H_{1}) + \epsilon}$, $ \frac{H_{2} - \text{mean}(H_{2})}{\text{std}(H_{2}) + \epsilon}$\Comment{Normalization}
			
			\State $C \gets H_{1}^T H_{2}$ \Comment{Compute cross-correlation}
			
			\State $\mathcal{L}_{DCCA}(H_{1}, H_{2}) \gets -\frac{\text{Tr}(C)}{\|H_{1}\|_F \|H_{2}\|_F + \epsilon}$\Comment{Deep CCA}
			
			\State \textbf{return} $\mathcal{L}_{DCCA}(H_{1}, H_{2})$
		\end{algorithmic}
		\label{dcca}
	\end{algorithm}
	\section{Experiments}\label{exp}
	
	\subsection{Datasets}
	We retrospectively collected a dataset of NCCT images stratified into four grades of EV severity, based on established classification systems~\citep{dahong2000trial, kawano2008short, wan2022quantitative}. The grading system considers two key factors: the form (F) of varices and the presence of red color sign (RC), with F1 (straight varices), F2 (enlarged and tortuous), and F3 (very large). The grades are defined as:
	$G0$: No detectable varices;
	$G1$ (\textit{Mild}): Varices of form F1 with no red color sign (RC-);
	$G2$ (\textit{Moderate}): Includes varices either with form F1 and a red color sign (RC+), or form F2 without a red color sign (RC-);
	$G3$ (\textit{Severe}): Varices characterized by form F2 with a red color sign (RC+), or form F3 with or without a red color sign (RC+ or RC-).
	For establishing the reference standard, we implemented a multi-step process:  endoscopic images were independently assessed by two gastroenterologists, with any discrepancies resolved by a senior gastroenterologist ($\geq$10 years of experience).

\begin{figure}[ht!]
		\centering
		\centering
		\includegraphics[width=1\linewidth]{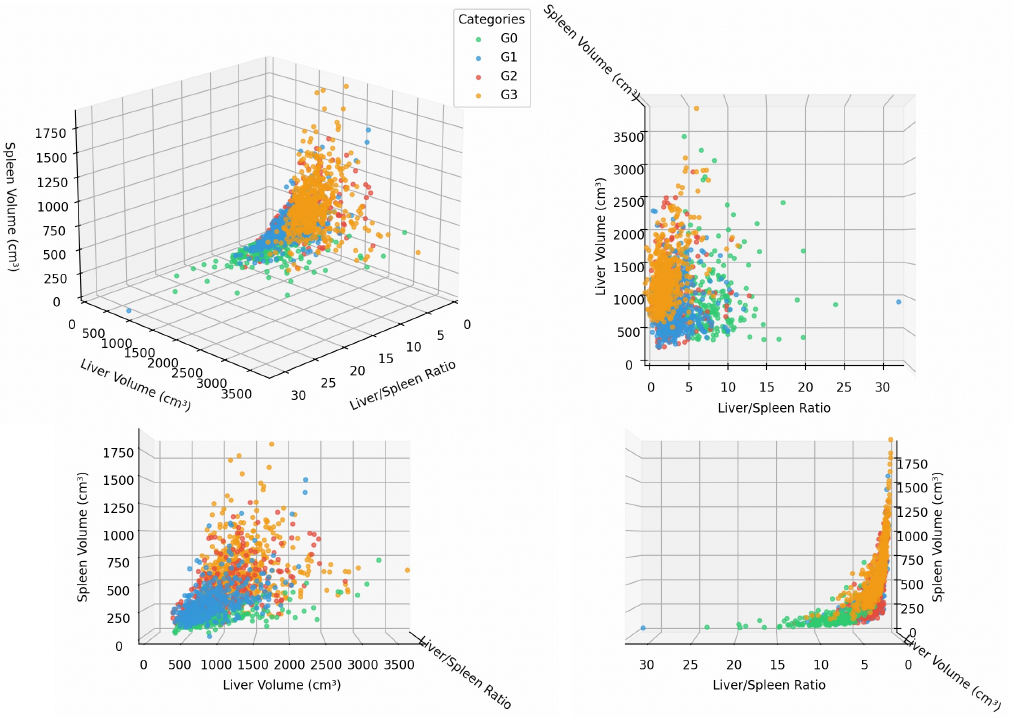}
		\caption{3D visualization and projection of the liver, spleen volume and LSVR.}
		\label{fig:3d}
	\end{figure}

	This dataset comprises CT images of 1,474 patients who underwent endoscopic examination for EV as shown in Table~\ref{dataset}.  Chest CT scans were acquired with Philips Ingenuity 4 and Siemens Sensation 64 CT scanners within one month before endoscopy. Images were standardized through resampling to isotropic 1mm spacing and applying abdominal window settings (width 400 HU, level 50 HU). From the total cohort, we randomly selected 239 subjects for the validation set and 289 subjects for the independent test cohort, maintaining the original grade distribution proportions. We included patients aged 18-85 years from 2018-2022, excluding those with prior esophageal interventions before CT examination.
	\begin{table}[ht!]\caption{Clinical characteristics and CT imaging parameters of the internal and external validation cohorts. MASLD: Metabolic dysfunction-Associated Steatotic Liver Disease; CHOL: Cholestatic liver disease; ALD: Alcoholic Liver Disease.}
		\resizebox{\columnwidth}{!}{
			\begin{tabular}{llll}
				\hline
				Characteristic                                          &               & Internal (n=1,474) & External (n=157) \\ \hline
				\multicolumn{2}{l}{Age (years), mean $\pm$Std.}                         & 54.19 $\pm$10.78   & 57.39 $\pm$10.16 \\
				\multirow{2}{*}{\textbackslash{}colrule Gender, n (\%)} & Male          & 986 (66.9\%)       & 97(61.8\%)       \\
				& Female        & 488 (33.1\%)       & 60(38.2\%)       \\ \hline
				\multirow{4}{*}{EV Staging, n (\%)}                     & G0            & 218 (14.8\%)       & 23 (14.6\%)      \\
				& G1            & 410 (27.8\%)       & 43 (27.3\%)      \\
				& G2            & 324 (22.0\%)       & 37 (23.6\%)      \\
				& G3            & 522 (35.4\%)       & 54 (34.4\%)      \\ \hline
				\multirow{5}{*}{Etiology, n (\%)}                       & Hepatitis B/C & 799 (64.0\%)       & 68 (43.3\%)      \\
				& MASLD         & 112 (9.0\%)        & 23 (14.7\%)      \\
				& CHOL          & 125 (10.0\%)       & 25 (15.9\%)      \\
				& ALD+Viral     & 61 (4.9\%)         & 12 (7.6\%)       \\
				& Other         & 377 (25.6\%)       & 29 (18.5\%)      \\ \hline
				Slice Thickness (mm), range                             &               & 1-5                & 1-5              \\ \hline
				Tube Voltage (KVP), range                               &               & 100-140            & 120              \\ \hline
				\end{tabular}}
		\label{dataset}
	\end{table}
	Our analysis of organ volumes and LSVR revealed distinct patterns in different grades of EV. The 3D visualization in Figure~\ref{fig:3d} further shows that, although individual organ volumes show considerable overlap between grades, LSVR provides more reliable grade differentiation, e.g., a lower LSVR indicates swollen spleens due to portal hypertension or a contracted liver, suggesting more severe liver fibrosis. As shown in Table~\ref{tab:model}, using only volume information with machine learning approaches yields comparable performance in staging accuracy. We also established a level classification for these parameters using z-score thresholds derived from statistical analysis (Table~\ref{level}).
	
	\begin{table}[ht!]\caption{Volume ranges for organs and LSVR categorized into five descriptive levels.}
		\resizebox{\columnwidth}{!}{
			\begin{tabular}{l|l|l|l|l|l}
				\hline
				Parameters      & Very Low & Low & Average    & High & Very High \\ \hline
				Eso. (cm³) & $\leq 15$      & $15-26$   & $26-50$    & $50-74$     & $\geq 74$       \\
				Liver (cm3)     & $\leq 297$     & $297-782$ & $782-1267$ & $1267-1753$ & $\geq 1753$     \\
				Spleen (cm³)    & $\leq 249$     & $249-389$ & $389-529$  & $529-809$   & $\geq 809$      \\
				LSVR            & $\leq 0.7$     & $0.7-2.4$ & $2.4-6.1$  & $6.1-7.9$  & $\geq 7.9$     \\ \hline
		\end{tabular}}
		\label{level}
	\end{table}
	\subsection{Implementation Details} 
	
	\begin{table}[ht!]\caption{Comparison of fine-tune nnUNet and Totalsegmentator on the segmentation test dataset using DICE coefficients.}
		\resizebox{\columnwidth}{!}{
			\begin{tabular}	{l|c|c|c|l}
				\hline
				\multicolumn{1}{c|}{Methods} & Esophagus & Liver & Spleen & Mean  \\ \hline
				TotalSegmentator~\citep{wasserthal2023totalsegmentator}             & 0.761     & 0.952 & 0.948  & 0.887 \\
				Fine-tune nnUNet~\citep{isensee2021nnu}             & 0.895     & 0.985 & 0.986  & 0.956 \\ \hline
		\end{tabular}}
		\label{tab:pretrain}
	\end{table}
	\begin{figure}[ht!]
		\centering
		\centering
		\includegraphics[width=1\linewidth]{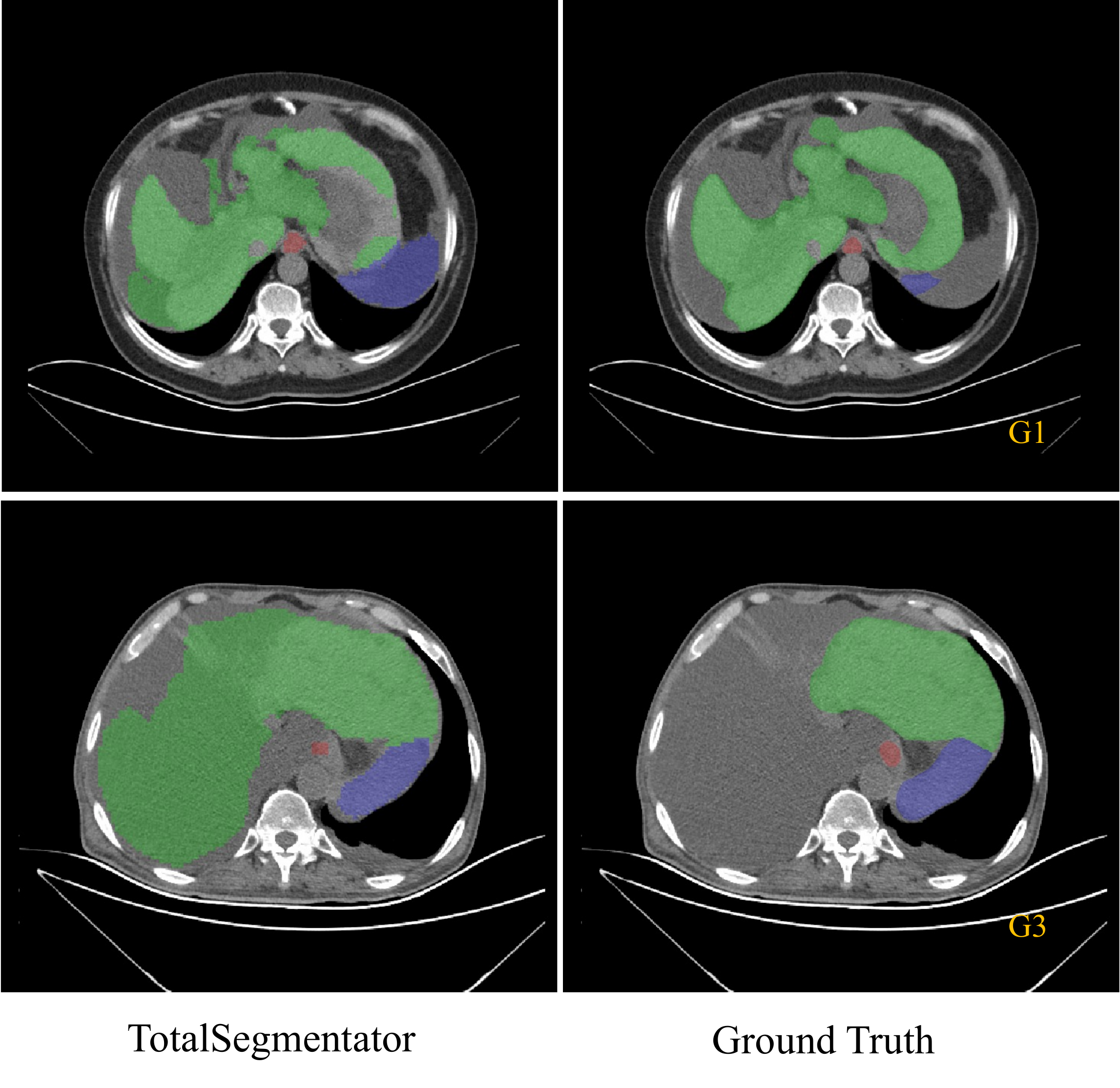}
		\caption{Challenges exist when using well pre-trained models, e.g., TotalSegmentator~\citep{wasserthal2023totalsegmentator} for organ segmentation, as multiple failure cases underscore the difficulty in generating accurate liver and spleen masks in subjects with liver cirrhosis.}
		\label{fig:seg_case}
	\end{figure}

	Liver and spleen segmentation in patients with liver cirrhosis presents significant challenges, particularly in decompensated cases, as illustrated in Figure~\ref{fig:seg_case}. In the conference version~\citep{li2024improved}, we initially fine-tuned the pre-trained Totalsegmentator~\citep{wasserthal2023totalsegmentator} (an nnUNet-based model~\citep{isensee2021nnu}) for accurate localization of the esophagus, liver, and spleen. To establish a high-quality reference standard for volume estimation, we adopted a hybrid approach combining automated segmentation with manual refinement. The model was first trained on 450 manually annotated masks randomly sampled from the EV training set, resulting in improved performance as shown in Table~\ref{tab:pretrain}. We employed the pre-trained nnUNet, configured for low-resolution image processing, as our localization network. Subsequently, all automated segmentation masks of the three organs were reviewed and refined by radiologists. The organ volumes were then calculated using  Equation~(\ref{volume}), and standardized text descriptions, including the LSVR, were generated for each organ.
	
	During training, we implemented various data augmentation techniques, including random rescaling, flipping, and cutout operations. The localized ROIs were subsequently resized to specific dimensions: esophagus ($40\times40\times100$), liver ($256\times196\times36$), and spleen ($152\times196\times24$). For image prior embedding, all full chest CT volumes were resized to $256\times256\times36$. Each branch of MOON++ utilized the UniFormer-base model~\citep{li2023uniformer}, pre-trained on Kinetics-400~\citep{kay2017kinetics} dataset. This pre-training approach benefits volumetric image analysis through learned spatio-temporal features~\citep{ke2024video}, producing feature representations with 512 channels. Multi-organ image backbone comprises $T=4$ downsampling blocks, and the ORI module operates with $N=8$ interaction steps to facilitate inter-organ feature exchange.
	
	The MOON++ framework was optimized using the Adam optimizer~\citep{kingma2014adam} with an initial learning rate of $10^{-5}$. Training proceeded for 100 epochs, with the learning rate decreasing by half every 20 epochs following a piecewise constant decay schedule. We set the weighting parameter $\lambda=0.9$ for Equation~(\ref{lcca}). The implementation was carried out in PyTorch and was executed on Nvidia A800 GPUs. For $f_1$ and $f_2$ in $\mathcal{L}_{DCCA}$, we use two simple fully-connected layers.
	\subsection{Main Experimental Results}
	Table~\ref{tab:main} presents a comprehensive evaluation of progressive enhancements to our EV grading methodology, with multi-class accuracy measuring exact matches across all four grades (G0-G3). The baseline single-organ model focusing solely on esophageal image input achieves reasonable performance, with AUC values of 0.870 for G3 classification in the test set, but shows limited multi-class accuracy (53.3\%). Adding multi-organ image features notably improves G1 detection, raising the AUC from 0.844 to 0.982. Incorporating the DCCA and ORI modules progressively boosts performance across all grades. Implementing the CLIP encoding image prior increases multi-class accuracy from 62.6\% to 64.7\% and Kendall's tau from 63.9\% to 71.4\% in the test set. The final enhancement of incorporating clinical textual features with CLIP encoding text prior yields the best overall performance, achieving the highest validation metrics (Multi. ACC: 69.8\%, $T_K$: 76.1\%) and test metrics (Multi. ACC: 65.3\%, $T_K$: 74.4\%). This progressive improvement demonstrates the effectiveness of each component in enhancing the MOON++ to accurately grade EV severity while maintaining ordinal relationships between grades.

	\begin{table*}[ht!]
		\caption{Comparative analysis between single-organ and multi-organ methodologies, examining various strategies for feature fusion. ACC(\%): Accuracy Values; AUC: Area under Curve Values. Multi. ACC(\%): Multi-class Accuracy. $T_K$(\%): Kendall's tau coefficient. }
		\resizebox{\textwidth}{!}{
			\begin{tabular}{l|cccc|cccc}
				\hline
				\multicolumn{1}{c|}{\multirow{3}{*}{Methods}} & \multicolumn{4}{c|}{Validation ($n=239$)}                                                                                                                              & \multicolumn{4}{c}{Independent Test ($n=289$)}                                                                                                                        \\ \cline{2-9} 
				\multicolumn{1}{c|}{}                         & \multicolumn{1}{c|}{$\geq$G1}        & \multicolumn{1}{c|}{$\geq$G2}        & \multicolumn{1}{c|}{G3}               & \multirow{2}{*}{Multi. ACC/$T_K$} & \multicolumn{1}{c|}{$\geq$G1}        & \multicolumn{1}{c|}{$\geq$G2}        & \multicolumn{1}{c|}{G3}               & \multirow{2}{*}{Multi. ACC/$T_K$} \\ \cline{2-4} \cline{6-8} 
				\multicolumn{1}{c|}{}                         & \multicolumn{1}{c|}{ACC / AUC}       & \multicolumn{1}{c|}{ACC / AUC}       & \multicolumn{1}{c|}{ACC / AUC}        &                                   & \multicolumn{1}{c|}{ACC / AUC}       & \multicolumn{1}{c|}{ACC / AUC}       & \multicolumn{1}{c|}{ACC / AUC}        &                                   \\ \hline
				Single-Organ (eso.)                           & \multicolumn{1}{c|}{$85.3/0.836$}          & \multicolumn{1}{c|}{$74.5/0.829$}          & \multicolumn{1}{c|}{$77.0/0.865$}          & $49.0/53.7$                       & \multicolumn{1}{c|}{$88.6/0.844$}          & \multicolumn{1}{c|}{$74.7/0.850$}          & \multicolumn{1}{c|}{$79.6/0.870$}          & $53.3/60.1$                       \\
				+Multi-Organs                                 & \multicolumn{1}{c|}{$91.6/0.968$}          & \multicolumn{1}{c|}{$77.8/0.824$}          & \multicolumn{1}{c|}{$76.9/0.845$}          & $57.7/60.9$                       & \multicolumn{1}{c|}{$92.7/0.982$}          & \multicolumn{1}{c|}{$77.9/0.858$}          & \multicolumn{1}{c|}{$78.5/0.857$}          & $56.1/63.4$                       \\
				+ORI                                          & \multicolumn{1}{c|}{\textbf{93.7}/\textbf{0.974}} & \multicolumn{1}{c|}{$76.1/0.825$}          & \multicolumn{1}{c|}{$80.3/0.854$}          & $58.9/63.4$                       & \multicolumn{1}{c|}{$93.8/\textbf{0.986}$}    & \multicolumn{1}{c|}{$81.0/0.876$}          & \multicolumn{1}{c|}{$79.2/0.876$}          & $61.6/63.8$                       \\
				+DCCA                                         & \multicolumn{1}{c|}{\textbf{93.7}/0.959}    & \multicolumn{1}{c|}{$80.3/0.856$}          & \multicolumn{1}{c|}{$76.5/0.856$}          & $60.3/63.5$                       & \multicolumn{1}{c|}{$95.5/0.985$}          & \multicolumn{1}{c|}{$76.8/0.862$}          & \multicolumn{1}{c|}{\textbf{80.6}/0.861}    & $62.6/63.9$                       \\
				+CLIP (img)                                   & \multicolumn{1}{c|}{$92.0/0.952$}          & \multicolumn{1}{c|}{$81.2/0.891$}          & \multicolumn{1}{c|}{$89.1/0.923$}          & $65.3/70.2$                       & \multicolumn{1}{c|}{$95.7/0.981$}          & \multicolumn{1}{c|}{$81.9/0.879$}          & \multicolumn{1}{c|}{$80.0/0.876$}          & $64.7/71.4$                       \\
				+CLIP (img+text)                              & \multicolumn{1}{c|}{$92.8/0.962$}          & \multicolumn{1}{c|}{\textbf{82.0}/\textbf{0.900}} & \multicolumn{1}{c|}{\textbf{91.2}/\textbf{0.959}} & \textbf{69.8}/\textbf{76.1}         & \multicolumn{1}{c|}{\textbf{96.1}/\textbf{0.986}} & \multicolumn{1}{c|}{\textbf{84.0}/\textbf{0.920}} & \multicolumn{1}{c|}{$80.2/\textbf{0.894}$}    & \textbf{65.3}/\textbf{74.4}         \\ \hline
		\end{tabular}}
		\label{tab:main}
	\end{table*}
	
	\subsection{Ablations Study}
	\noindent\textbf{Backbones.} 
	Table~\ref{tab:model} presents a comprehensive ablation study comparing different deep learning architectures in various organ inputs and radiologists' performance. ResNet3D-18 performs robustly in single-organ analysis, achieving 94.5\% accuracy for G1 detection in the liver and spleen on the test set, indicating its efficacy in capturing local features. However, UniFormer-B, which integrates CNN and transformer strengths, excels with full scan inputs, achieving 96.2\% accuracy and an AUC of 0.988 for G1 detection. The organ-specific analysis reveals distinct strengths: esophageal scans excel in G3 detection (highest ACC: 79.6\%, AUC: 0.870 with UniFormer-B), liver scans show strong G2 detection (best ACC: 79.1\% in validation), while spleen scans demonstrate high G1 detection but lower G3 performance. 
	
	Single organ based approaches achieve performance comparable to radiologists' assessments (48.4\% and 52.2\% multi-class accuracy), with UniFormer-B using esophageal input reaching 53.3\% multi-class accuracy. The comparison between full-scan and single-organ approaches shows task-dependent patterns: full scan configurations excel in G1 detection, while single-organ approaches, particularly esophageal scans, show superior G3 detection performance. These findings indicate that combining multi-organ and single-organ strengths within the UniFormer framework optimizes EV grading, as supported by the main results. 
	
\begin{figure}[ht!]
		\centering
		\centering
		\includegraphics[width=1\linewidth]{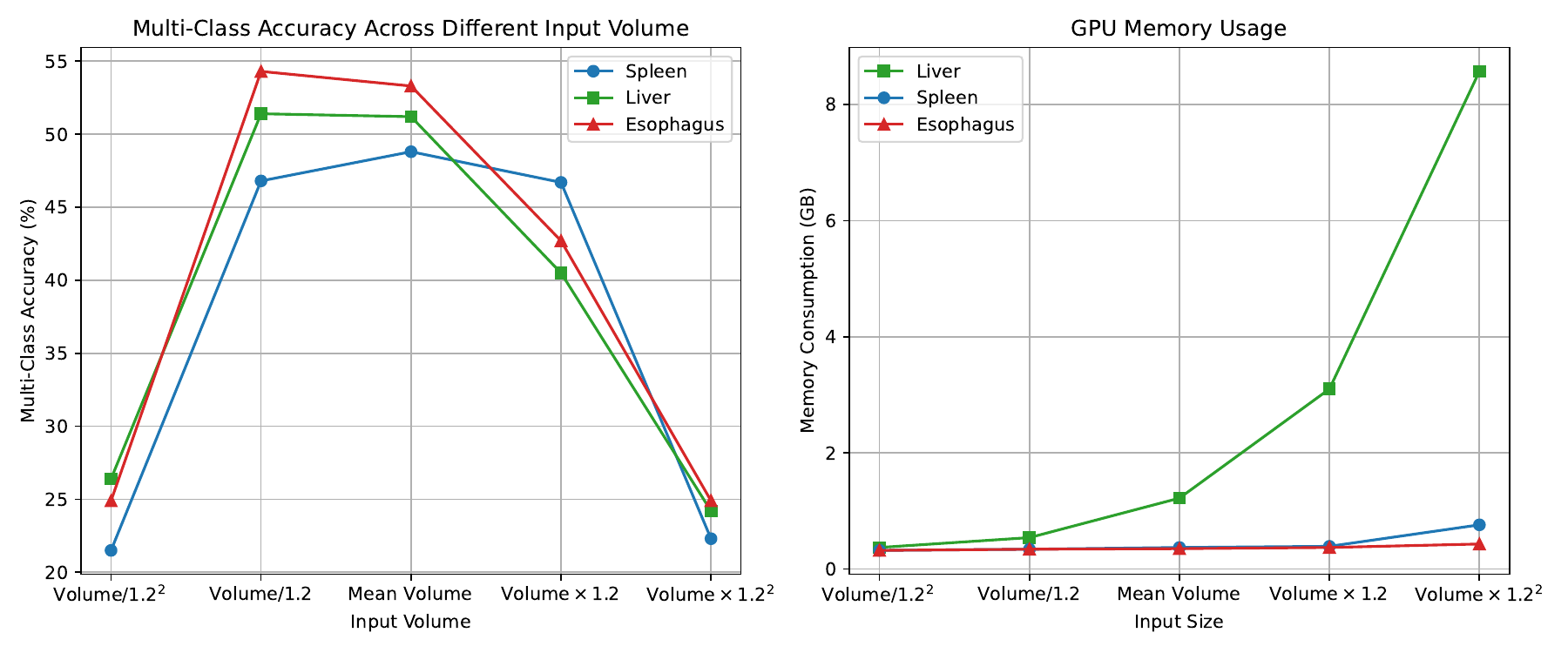}
		\caption{Comparison of different input volumes using UniFormer. The mean volumes of the esophagus, liver, and spleen are 60$\times$76$\times$110, 297$\times$259$\times$37, and 153$\times$191$\times$72, respectively.}
		\label{fig:input}
	\end{figure}

	As shown in Figure~\ref{fig:input}, the size of the input volume significantly affects performance. Optimal results are achieved with input volumes slightly smaller than or equal to the average cropped size (detailed in Section 4.2). Larger input volumes not only degrade performance but also substantially increase memory requirements.

\begin{figure}[ht!]
		\centering
		\centering
		\includegraphics[width=1\linewidth]{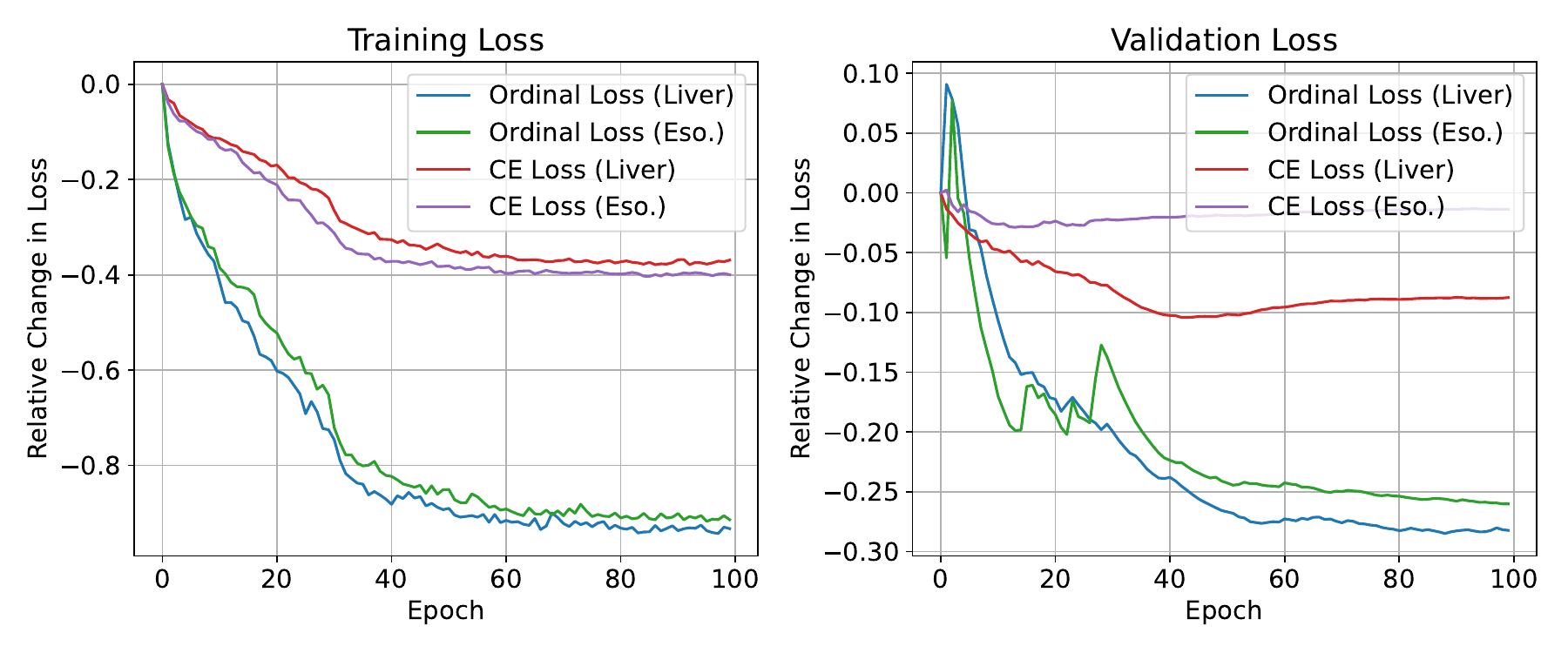}
		\caption{Relative loss curves for different loss functions.}
		\label{fig:loss}
	\end{figure}

	\begin{table*}[ht!]
		\caption{Comparative analysis of various organ inputs using different methods. Models include machine learning methods, ResNet3D-18~\citep{he2016deep}, SwinTransformer-B~\citep{liu2021swin}, and UniFormer-B~\citep{li2023uniformer}, all pretrained on Kinetics-400~\citep{kay2017kinetics}. ACC(\%): Accuracy Values; AUC: Area under Curve Values. Multi. ACC(\%): Multi-class Accuracy. }
		\resizebox{\textwidth}{!}{
			\begin{tabular}{l|l|cccc|cccc}
				\hline
				\multirow{3}{*}{Inputs} & \multirow{3}{*}{Methods} & \multicolumn{4}{c|}{Validation ($n=239$)} & \multicolumn{4}{c}{Independent Test ($n=289$)} \\ \cline{3-10} 
				& & \multicolumn{1}{c|}{$\geq$G1} & \multicolumn{1}{c|}{$\geq$G2} & \multicolumn{1}{c|}{G3} & \multirow{2}{*}{Multi. ACC} & \multicolumn{1}{c|}{$\geq$G1} & \multicolumn{1}{c|}{$\geq$G2} & \multicolumn{1}{c|}{G3} & \multirow{2}{*}{Multi. ACC} \\ \cline{3-5} \cline{7-9} 
				& & \multicolumn{1}{c|}{ACC / AUC} & \multicolumn{1}{c|}{ACC / AUC} & \multicolumn{1}{c|}{ACC / AUC} & & \multicolumn{1}{c|}{ACC / AUC} & \multicolumn{1}{c|}{ACC / AUC} & \multicolumn{1}{c|}{ACC / AUC} & \\ \hline
				Full Scan & Radiologist1 & \multicolumn{1}{c|}{-} & \multicolumn{1}{c|}{-} & \multicolumn{1}{c|}{-} & - & \multicolumn{1}{c|}{$72.7/0.837$} & \multicolumn{1}{c|}{$71.2/0.709$} & \multicolumn{1}{c|}{$74.6/0.796$} & $48.4$ \\
				Full Scan & Radiologist2 & \multicolumn{1}{c|}{-} & \multicolumn{1}{c|}{-} & \multicolumn{1}{c|}{-} & - & \multicolumn{1}{c|}{$75.2/0.848$} & \multicolumn{1}{c|}{$74.2/0.734$} & \multicolumn{1}{c|}{$74.8/0.806$} & $52.2$ \\ \hline
				Volumes+LSVR & Random Forest & \multicolumn{1}{c|}{$86.1/0.916$} & \multicolumn{1}{c|}{$70.2/0.791$} & \multicolumn{1}{c|}{\textbf{80.7}/0.886} & $51.4$ & \multicolumn{1}{c|}{$90.6/0.958$} & \multicolumn{1}{c|}{$73.0/0.828$} & \multicolumn{1}{c|}{$66.7 / 0.764$} & $49.8$ \\
				Volumes+LSVR & SVM & \multicolumn{1}{c|}{$88.2/0.874$} & \multicolumn{1}{c|}{$70.7/0.778$} & \multicolumn{1}{c|}{$74.4/0.837$} & $53.1$ & \multicolumn{1}{c|}{$88.5 / 0.883$} & \multicolumn{1}{c|}{$77.1/0.841$} & \multicolumn{1}{c|}{$60.5/0.767$} & $49.4$ \\
				Volumes+LSVR & Logistic Regression & \multicolumn{1}{c|}{$88.7/0.876$} & \multicolumn{1}{c|}{$63.6/0.773$} & \multicolumn{1}{c|}{$54.8/0.857$} & $45.6$ & \multicolumn{1}{c|}{$90.3/0.911$} & \multicolumn{1}{c|}{$69.2/0.812$} & \multicolumn{1}{c|}{$56.4/0.829$} & $48.8$ \\ \hline
				Full Scan & ResNet & \multicolumn{1}{c|}{$90.8 / 0.944$} & \multicolumn{1}{c|}{$75.3 / 0.791$} & \multicolumn{1}{c|}{$67.8 / 0.740$} & $44.8$ & \multicolumn{1}{c|}{\textbf{97.6}/0.747} & \multicolumn{1}{c|}{\textbf{78.9}/ 0.723} & \multicolumn{1}{c|}{$77.1 / 0.845$} & $51.2$ \\
				Single-organ (Liver) & ResNet & \multicolumn{1}{c|}{$92.9/0.972$} & \multicolumn{1}{c|}{$76.2 / 0.828$} & \multicolumn{1}{c|}{$72.8/\textbf{0.975}$} & $52.3$ & \multicolumn{1}{c|}{$94.5 / 0.978$} & \multicolumn{1}{c|}{$75.8 / 0.811$} & \multicolumn{1}{c|}{$74.4/0.811$} & $52.1$ \\
				Single-organ (Spleen) & ResNet & \multicolumn{1}{c|}{$93.3/0.958$} & \multicolumn{1}{c|}{$76.2 / 0.824$} & \multicolumn{1}{c|}{$68.6/0.773$} & $50.2$ & \multicolumn{1}{c|}{$94.5 / 0.974$} & \multicolumn{1}{c|}{$78.1 / 0.685$} & \multicolumn{1}{c|}{$74.3/0.833$} & $47.4$ \\
				Single-organ (Eso.) & ResNet & \multicolumn{1}{c|}{$86.2/0.867$} & \multicolumn{1}{c|}{$76.6/0.802$} & \multicolumn{1}{c|}{$73.2/0.821$} & $44.9$ & \multicolumn{1}{c|}{$87.9 / 0.822$} & \multicolumn{1}{c|}{$71.6/0.792$} & \multicolumn{1}{c|}{$75.1/0.810$} & $48.4$ \\ \hline
				Full Scan & SwinT. & \multicolumn{1}{c|}{$90.4 / 0.943$} & \multicolumn{1}{c|}{$73.2 / 0.751$} & \multicolumn{1}{c|}{$62.3 / 0.680$} & $41.0$ & \multicolumn{1}{c|}{$87.9 / 0.910$} & \multicolumn{1}{c|}{$70.2 / 0.767$} & \multicolumn{1}{c|}{$70.2 / 0.730$} & $39.4$ \\
				Single-organ (Liver) & SwinT. & \multicolumn{1}{c|}{$89.1 / 0.916$} & \multicolumn{1}{c|}{$70.7 / 0.724$} & \multicolumn{1}{c|}{$66.5 / 0.705$} & $43.1$ & \multicolumn{1}{c|}{$89.6 / 0.916$} & \multicolumn{1}{c|}{$72.7 / 0.765$} & \multicolumn{1}{c|}{$67.1 / 0.737$} & $43.6$ \\
				Single-organ (Spleen) & SwinT. & \multicolumn{1}{c|}{$91.6 / 0.965$} & \multicolumn{1}{c|}{$78.7/ 0.808$} & \multicolumn{1}{c|}{$66.5 / 0.752$} & $51.0$ & \multicolumn{1}{c|}{$93.8 / 0.975$} & \multicolumn{1}{c|}{$74.0 / 0.793$} & \multicolumn{1}{c|}{$69.6 / 0.753$} & $48.1$ \\
				Single-organ (Eso.) & SwinT. & \multicolumn{1}{c|}{$86.6 / 0.858$} & \multicolumn{1}{c|}{$75.7 / 0.817$} & \multicolumn{1}{c|}{$74.9 / 0.835$} & $50.6$ & \multicolumn{1}{c|}{$84.0 / 0.822$} & \multicolumn{1}{c|}{$75.4 / 0.837$} & \multicolumn{1}{c|}{$78.5 / 0.859$} & $49.5$ \\ \hline
				Full Scan & UniFormer & \multicolumn{1}{c|}{\textbf{94.6}/\textbf{0.978}} & \multicolumn{1}{c|}{$77.4 / 0.849$} & \multicolumn{1}{c|}{$75.3 / 0.798$} & $53.6$ & \multicolumn{1}{c|}{$96.2/\textbf{0.988}$} & \multicolumn{1}{c|}{$73.1 / 0.777$} & \multicolumn{1}{c|}{$74.8 / 0.768$} & $51.7$ \\
				Single-organ (Liver) & UniFormer & \multicolumn{1}{c|}{$93.7 / 0.976$} & \multicolumn{1}{c|}{\textbf{79.1}/\textbf{0.859}} & \multicolumn{1}{c|}{$75.3 / 0.840$} & \textbf{56.9} & \multicolumn{1}{c|}{$94.8 / 0.976$} & \multicolumn{1}{c|}{$78.2 / 0.812$} & \multicolumn{1}{c|}{$72.3 / 0.790$} & $51.2$ \\
				Single-organ (Spleen) & UniFormer & \multicolumn{1}{c|}{$93.3 / 0.964$} & \multicolumn{1}{c|}{$78.2 / 0.835$} & \multicolumn{1}{c|}{$66.9 / 0.761$} & $50.2$ & \multicolumn{1}{c|}{$95.2 / 0.970$} & \multicolumn{1}{c|}{$72.7 / 0.779$} & \multicolumn{1}{c|}{$68.9 / 0.742$} & $48.8$ \\
				Single-organ (Eso.) & UniFormer & \multicolumn{1}{c|}{$85.3/ 0.836$} & \multicolumn{1}{c|}{$74.5 / 0.829$} & \multicolumn{1}{c|}{$77.0/0.865$} & $49.0$ & \multicolumn{1}{c|}{$88.6 / 0.844$} & \multicolumn{1}{c|}{$74.7 / \textbf{0.850}$} & \multicolumn{1}{c|}{\textbf{79.6}/\textbf{0.870}} & \textbf{53.3} \\ \hline
		\end{tabular}}
		\label{tab:model}
	\end{table*}
	\noindent\textbf{Loss Functions.} 
	We evaluated different loss functions using liver and esophagus baselines, as they showed the highest baseline performance (Table~\ref{tab:model}), with results presented in Table~\ref{loss} and Figure~\ref{fig:loss}. Ordinal regression loss demonstrated superior performance over cross-entropy (CE) and hybrid approaches in most test scenarios. In liver-based evaluation, ordinal regression showed particular strength in early-stage detection ($\geq$G1), significantly outperforming conventional CE approaches. For severe cases (G3), the ordinal regression maintained a robust discrimination capability. This superior performance is due to the ability of ordinal regression to capture the progressive nature of EV severity (G0 $\textless$ G1 $\textless$ G2 $\textless$ G3) and maintain stable training dynamics. For $\mathcal{L}_{\text{DCCA}}$, the weighting parameter $\lambda$ was set to 0.9 based on a sensitivity analysis shown in Table~\ref{tab:lambda}.

	\begin{table*}
		\caption{Comparison analysis of different loss functions for EV grading.  ACC(\%): Accuracy Values; AUC: Area under Curve Values. Multi. ACC(\%): Multi-class Accuracy. $T_K$(\%): Kendall's tau coefficient. Note that CE implementation requires ReLU activation in the classifier layer.}
		\resizebox{\textwidth}{!}{
			\begin{tabular}{c|c|l|cccc|cccc}
				\hline
				\multirow{3}{*}{Organs}               & \multirow{3}{*}{Methods}   & \multirow{3}{*}{Loss Function} & \multicolumn{4}{c|}{Validation ($n=239$)}                                                                                                                                      & \multicolumn{4}{c}{Independent Test ($n=289$)}                                                                                                                                 \\ \cline{4-11} 
				&                            &                                & \multicolumn{1}{c|}{$\geq$G1}                & \multicolumn{1}{c|}{$\geq$G2}                & \multicolumn{1}{c|}{G3}                      & \multirow{2}{*}{Multi. ACC/$T_K$} & \multicolumn{1}{c|}{$\geq$G1}                & \multicolumn{1}{c|}{$\geq$G2}                & \multicolumn{1}{c|}{G3}                      & \multirow{2}{*}{Multi. ACC/$T_K$} \\ \cline{4-6} \cline{8-10}
				&                            &                                & \multicolumn{1}{c|}{ACC / AUC}               & \multicolumn{1}{c|}{ACC / AUC}               & \multicolumn{1}{c|}{ACC / AUC}               &                                   & \multicolumn{1}{c|}{ACC / AUC}               & \multicolumn{1}{c|}{ACC / AUC}               & \multicolumn{1}{c|}{ACC / AUC}               &                                   \\ \hline
				\multirow{3}{*}{Single-organ (Liver)} & \multirow{3}{*}{UniFormer} & Ordinal                        & \multicolumn{1}{c|}{$\textbf{93.7/0.976}$} & \multicolumn{1}{c|}{$\textbf{79.1/0.859}$} & \multicolumn{1}{c|}{$75.3 / 0.840$}          & $\textbf{56.9/60.5}$              & \multicolumn{1}{c|}{$94.8 / \textbf{0.976}$} & \multicolumn{1}{c|}{$\textbf{78.2} / 0.812$} & \multicolumn{1}{c|}{$72.3 / 0.790$}          & $51.2/58.7$                       \\
				&                            & CE                             & \multicolumn{1}{c|}{$93.7 / 0.983$}          & \multicolumn{1}{c|}{$76.9 / 0.851$}          & \multicolumn{1}{c|}{$71.3 / 0.802$}          & $56.4/57.6$                       & \multicolumn{1}{c|}{$\textbf{95.5} / 0.968$}          & \multicolumn{1}{c|}{$74.0 / 0.819$}          & \multicolumn{1}{c|}{$71.3 / 0.782$}          & $52.4/55.7$                       \\
				&                            & 0.5CE+0.5Ordinal               & \multicolumn{1}{c|}{$76.5 / 0.845$}          & \multicolumn{1}{c|}{$66.5 / 0.778$}          & \multicolumn{1}{c|}{$68.2 / 0.809$}          & $44.4/33.1$                       & \multicolumn{1}{c|}{$76.5 / 0.922$}          & \multicolumn{1}{c|}{$63.0 / 0.752$}          & \multicolumn{1}{c|}{$65.3 / 0.828$}          & $41.9/30.8$                       \\ \hline
				\multirow{3}{*}{Single-organ (Eso.)}  & \multirow{3}{*}{UniFormer} & Ordinal                        & \multicolumn{1}{c|}{$85.3 / 0.836$}          & \multicolumn{1}{c|}{$74.5 / 0.829$}          & \multicolumn{1}{c|}{$\textbf{77.0/0.865}$} & $49.0/53.7$                       & \multicolumn{1}{c|}{$88.6 / 0.844$}          & \multicolumn{1}{c|}{$74.7 / \textbf{0.850}$} & \multicolumn{1}{c|}{$\textbf{79.6/0.870}$} & $\textbf{53.3/60.1}$              \\
				&                            & CE                             & \multicolumn{1}{c|}{$84.1/0.867$}            & \multicolumn{1}{c|}{$69.8/0.803$}            & \multicolumn{1}{c|}{$72.8/0.821$}            & $47.6/43.3$                       & \multicolumn{1}{c|}{$88.5 / 0.835$}          & \multicolumn{1}{c|}{$75.7 / 0.845$}          & \multicolumn{1}{c|}{$78.5 / 0.843$}          & $55.0/57.4$                       \\
				&                            & 0.5CE+0.5Ordinal               & \multicolumn{1}{c|}{$75.3 / 0.794$}          & \multicolumn{1}{c|}{$62.8 / 0.740$}          & \multicolumn{1}{c|}{$71.5 / 0.839$}          & $40.6/33.2$                       & \multicolumn{1}{c|}{$76.5 / 0.775$}          & \multicolumn{1}{c|}{$64.4 / 0.773$}          & \multicolumn{1}{c|}{$71.2 / 0.849$}          & $42.9/38.9$                       \\ \hline
		\end{tabular}}
		\label{loss}
	\end{table*}

	\begin{table}[ht!]
		\caption{Sensitivity analysis of the weighting parameter $\lambda$. ACC(\%): Accuracy Values; AUC: Area under Curve Values. Multi. ACC(\%): Multi-class Accuracy. $T_K$(\%): Kendall's tau coefficient.}
		\resizebox{\columnwidth}{!}{
			\begin{tabular}{l|c|cccc}
				\hline
				\multirow{3}{*}{Methods}      & \multirow{3}{*}{$\lambda$} & \multicolumn{4}{c}{Independent Test ($n=289$)}                                                                                                                                  \\ \cline{3-6} 
				&                            & \multicolumn{1}{c|}{$\geq$G1}              & \multicolumn{1}{c|}{$\geq$G2}                     & \multicolumn{1}{c|}{G3}                    & \multirow{2}{*}{Multi. ACC/$T_K$} \\ \cline{3-5}
				&                            & \multicolumn{1}{c|}{ACC / AUC}             & \multicolumn{1}{c|}{ACC / AUC}                    & \multicolumn{1}{c|}{ACC / AUC}             &                                   \\ \hline
				\multirow{5}{*}{Multi-Organs} & 1.00                          & \multicolumn{1}{c|}{$93.8/\textbf{0.986}$} & \multicolumn{1}{c|}{\textbf{81.0}/\textbf{0.876}} & \multicolumn{1}{c|}{$79.2/\textbf{0.876}$} & $61.6/63.8$                       \\
				& 0.95                       & \multicolumn{1}{c|}{$94.3/0.984$}          & \multicolumn{1}{c|}{$78.4/0.864$}                 & \multicolumn{1}{c|}{\textbf{81.2}/0.873}   & $62.1/63.9$                       \\
				& 0.90                        & \multicolumn{1}{c|}{\textbf{95.5}/0.985}   & \multicolumn{1}{c|}{$76.8/0.862$}                 & \multicolumn{1}{c|}{$80.6/0.861$}          & \textbf{62.6}/\textbf{63.9}       \\
				& 0.80                        & \multicolumn{1}{c|}{$95.1/0.982$}          & \multicolumn{1}{c|}{$76.9/0.859$}                 & \multicolumn{1}{c|}{$79.3/0.872$}          & $61.3/63.4$                       \\
				& 0.50                        & \multicolumn{1}{c|}{$94.2/0.985$} & \multicolumn{1}{c|}{$79.3/0.846$}                 & \multicolumn{1}{c|}{$78.4/0.856$}          & $59.3/63.1$                       \\ \hline
		\end{tabular}}
		\label{tab:lambda}
	\end{table}
	
	\begin{table}[ht!]
		\caption{Comparison of different feature interaction strategies, with details illustrated in Figure \ref{fig:inter}. We report the parameters and Floating Point Operations (FLOPs) for the first multi-organ image backbone. ACC(\%): Accuracy Values; AUC: Area under Curve Values. Multi. ACC(\%): Multi-class Accuracy. $T_K$(\%): Kendall's tau coefficient.}
		\resizebox{\columnwidth}{!}{
			\begin{tabular}{l|l|l|l|cccc}
				\hline
				\multirow{3}{*}{Feature Fusion} & \multirow{3}{*}{Complexity} & \multirow{3}{*}{Param. (M)} & \multirow{3}{*}{FLOPs (G)} & \multicolumn{4}{c}{Independent Test ($n=289$)}                                                                                                                               \\ \cline{5-8} 
				&                             &                             &                        & \multicolumn{1}{c|}{$\geq$G1}                & \multicolumn{1}{c|}{$\geq$G2}                & \multicolumn{1}{c|}{G3}                    & \multirow{2}{*}{Multi. ACC/$T_K$} \\ \cline{5-7}
				&                             &                             &                        & \multicolumn{1}{c|}{ACC / AUC}               & \multicolumn{1}{c|}{ACC / AUC}               & \multicolumn{1}{c|}{ACC / AUC}             &                                   \\ \hline
				-                               & -                           & -                           & -                      & \multicolumn{1}{c|}{$92.7 / 0.982$}          & \multicolumn{1}{c|}{$77.9 / 0.858$}          & \multicolumn{1}{c|}{$78.5 / 0.857$}        & $56.1 / 63.4$                     \\
				Add                             & $\mathcal{O}(1)$            & 0.000                       & 0.000                  & \multicolumn{1}{c|}{$95.1 / 0.987$}          & \multicolumn{1}{c|}{$79.2 / \textbf{0.885}$} & \multicolumn{1}{c|}{$78.2 / 0.867$}        & $58.1 / 62.5$                     \\
				Concat                          & $\mathcal{O}(1)$            & 0.012                       & 0.015                  & \multicolumn{1}{c|}{$90.7 / 0.981$}          & \multicolumn{1}{c|}{$80.6 / 0.881$}          & \multicolumn{1}{c|}{$73.0 / 0.873$}        & $55.3 / 64.2$                     \\
				(a)                             & $\mathcal{O}(4H W Z C^2)$   & 1.604                        & 1.639                   & \multicolumn{1}{c|}{$\textbf{95.8/0.986}$}   & \multicolumn{1}{c|}{$79.6 / 0.871$}          & \multicolumn{1}{c|}{$78.2 / 0.868$}        & $58.5 / 63.5$                     \\
				(b)                             & $\mathcal{O}(2H W Z C^2)$   & 0.819                        & 0.838                   & \multicolumn{1}{c|}{$94.8 / 0.987$}          & \multicolumn{1}{c|}{$78.2 / 0.851$}          & \multicolumn{1}{c|}{$77.2 / 0.865$}        & $59.5 / \textbf{64.5}$            \\
				(c)                             & $\mathcal{O}(H W Z C^2)$    & 0.937                        & 0.959                   & \multicolumn{1}{c|}{$93.8 / \textbf{0.986}$} & \multicolumn{1}{c|}{$\textbf{81.0} / 0.876$} & \multicolumn{1}{c|}{$\textbf{79.2/0.876}$} & $\textbf{61.6} / 63.8$            \\ \hline
		\end{tabular}}
		\label{tab:interaction}
	\end{table}

	\begin{figure}[ht!]
		\centering
		\centering
		\includegraphics[width=1\linewidth]{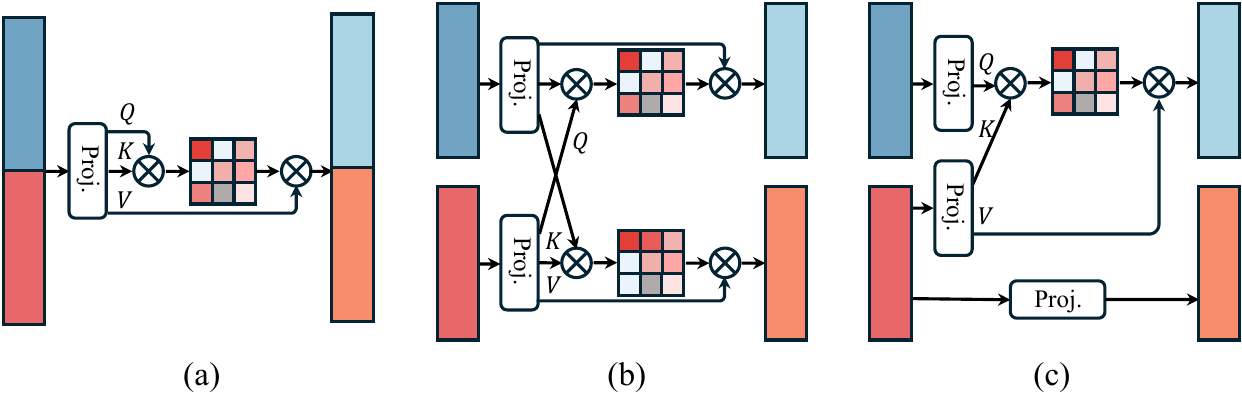}
		\caption{Different feature interaction strategies. (a) Self-attention on concatenated features. (b) Cross-attention with query swapping. (c) Cross-attention with a switching manner.}
		\label{fig:inter}
	\end{figure}
	
	\noindent\textbf{Feature Interaction.}  Different feature interaction strategies in the ORI module were evaluated, as illustrated in Figure~\ref{fig:inter} and quantified in Table~\ref{tab:interaction}. While simple fusion methods (Add, Concat) show moderate improvements over the baseline, attention-based interactions demonstrate superior performance. Strategy (c), employing cross-attention in a switching manner, achieves the best overall performance (61.6\% multi-class accuracy) while maintaining reasonable computational complexity ($\mathcal{O}(HWZ C^2)$). This approach balances effective feature interaction with computational efficiency, outperforming both self-attention on concatenated features (a) and query-swapped cross-attention (b), particularly in G2 and G3 detection (ACC/AUC: 81.0\%/0.876 and 79.2\%/0.876 respectively). These results suggest a selective feature interaction through switching cross-attention. 
	
	\noindent\textbf{Clinical Prior Embedding.} 
	Since the clinical priors are structured volumes and LSVR, we investigate optimal embedding methods for multi-modal learning, comparing simple one-hot encoding with medical CLIP encoders in Table~\ref{embed}. Our evaluation reveals that the addition of clinical priors consistently improves results across different embedding approaches.
	
	CLIP-like 3D image encoding demonstrates robust performance improvements (Multi. ACC: 64.7\%), benefiting from the rich pre-trained visual representations of the complete CT volumes. Text-only CLIP encoding exhibits lower performance (Multi. ACC: 61.2\%), likely due to noise introduced when processing structured clinical measurements through complex text encoders. Notably, simple one-hot encoding achieves comparable results (Multi. ACC: 63.1\%) and shows accuracy for G3 classification (ACC: 81.6\%), suggesting that relying solely on the CLIP text encoder may introduce more variability compared to one-hot encoding~\citep{Xu2024RemixFormerAM} when using structured input.
	
	While the full CLIP implementation combining image and text encoders achieves the best performance (Multi. ACC: 65.3\%), the modest improvement over simpler approaches (approximately 2.2\% accuracy gain compared to one-hot embed) suggests that both text and image encoders with CLIP may provide smoother embedding and better CT-text alignment.

	\noindent\textbf{CLIP Encoders.} 
	The analysis of dimensionality in Table \ref{clip_img} demonstrates that a 64-dimensional adaptor achieves optimal performance with a 64.7\% multi-class accuracy and a 71.4\% Kendall's tau. This optimal dimension suggests that, while CLIP's original high-dimensional space contains rich visual features, compressing it to 64 dimensions effectively distills EV-specific characteristics while filtering out potential noise. Note that performance degradation occurs with both larger and smaller dimensions, indicating a trade-off between feature preservation and model complexity.
	
	In the multi-modal investigation shown in Table \ref{clip_text}, balanced 64 + 64 dimensional adapters for image and text yield superior results, achieving a 65.3\% multi-class accuracy and 74.4\% Kendall's tau. This symmetric architecture likely facilitates an optimal shared semantic space, allowing visual and textual features to complement each other effectively. The decreased performance observed with unbalanced dimensions highlights the importance of modality alignment for effective feature fusion.
	
	Our ablation analysis in Table \ref{text_des} demonstrates that categorical descriptions outperformed numerical values in text representation, particularly when combining categorical volume descriptions with LSVR. This advantage of categorical encoding likely arises from its ability to capture clinically sensible/meaningful thresholds and relationships that raw numerical values might overlook. Therefore, we adopted the categorical text representation depicted in Figure~\ref{fig:pipe} for text input. The combination of categorical volume and LSVR provides the highest performance by leveraging complementary clinical indicators in a structured format.
	
	To investigate the necessity of additional image-text alignment, we conducted experiments with supplementary supervision methods, as shown in Table~\ref{supervision}. The results indicate that while methods like contrastive learning and InfoNCE~\citep{oord2018representation} provide slight enhancements in certain metrics, they do not significantly improve overall performance compared to the baseline. This suggests that the pre-trained representations already sufficiently maintain alignment patterns. Additionally, considering that effective contrastive learning typically requires large mini-batch sizes for stable training~\citep{radford2021learning}, our findings support maintaining model simplicity without additional alignment steps, offering practical benefits in training stability.
	
\begin{figure}[ht!]
		\centering
		\begin{subfigure}[b]{0.3\linewidth}
			\centering
			\includegraphics[width=\linewidth]{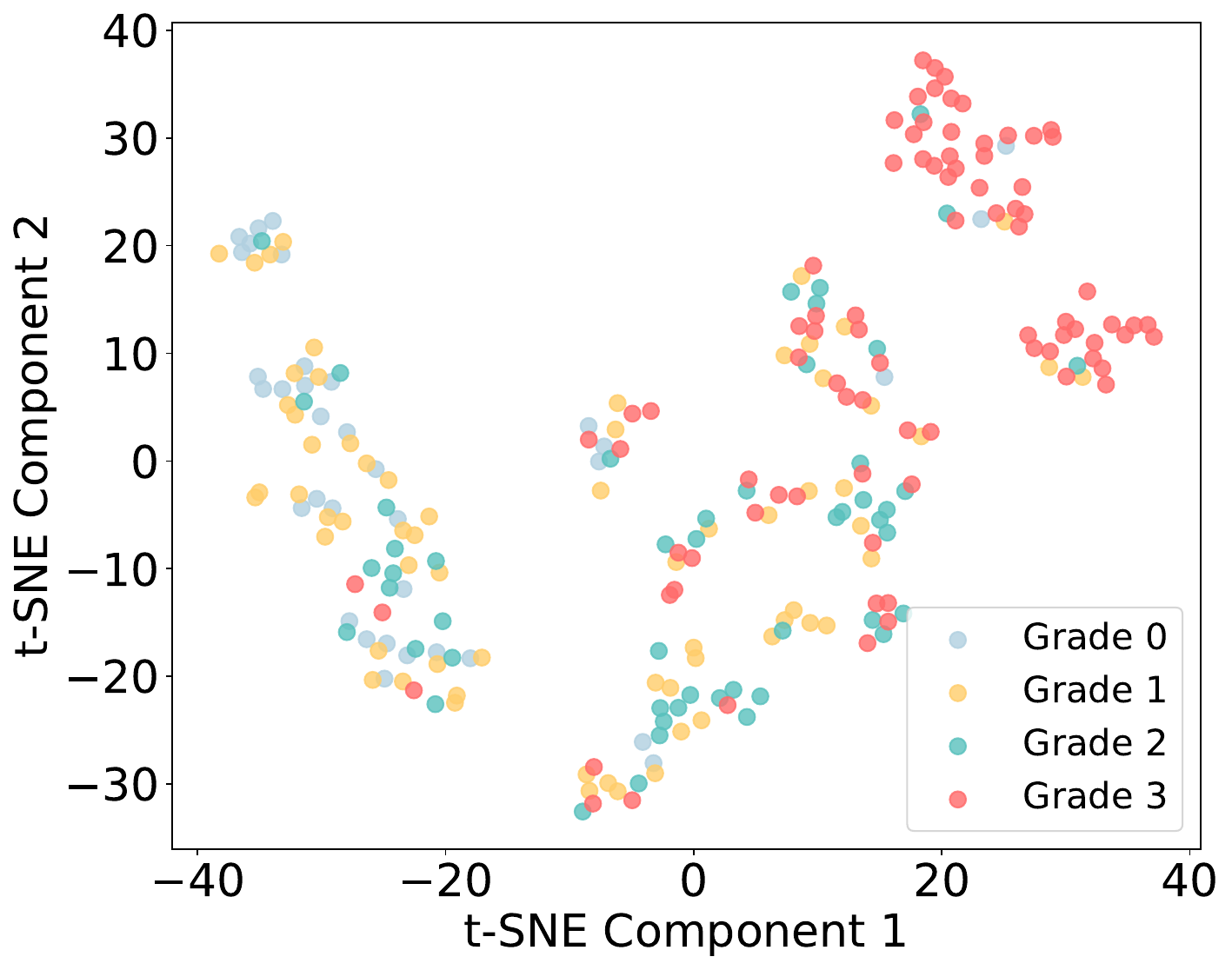}
			\label{fig:subim1}
		\end{subfigure}
		\hfill
		\begin{subfigure}[b]{0.3\linewidth}
			\centering
			\includegraphics[width=\linewidth]{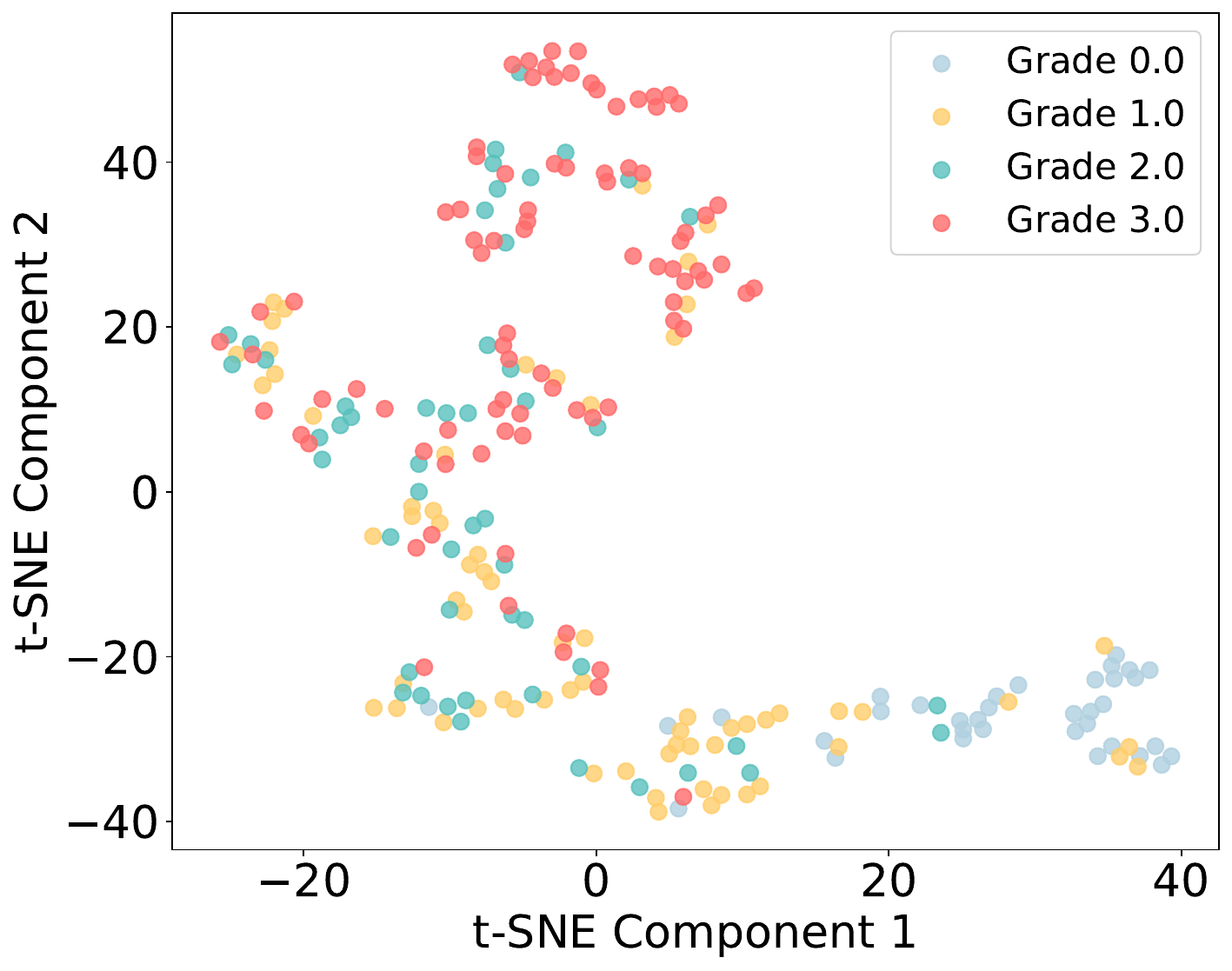}
			\label{fig:subim2}
		\end{subfigure}
		\hfill
		\begin{subfigure}[b]{0.3\linewidth}
			\centering
			\includegraphics[width=\linewidth]{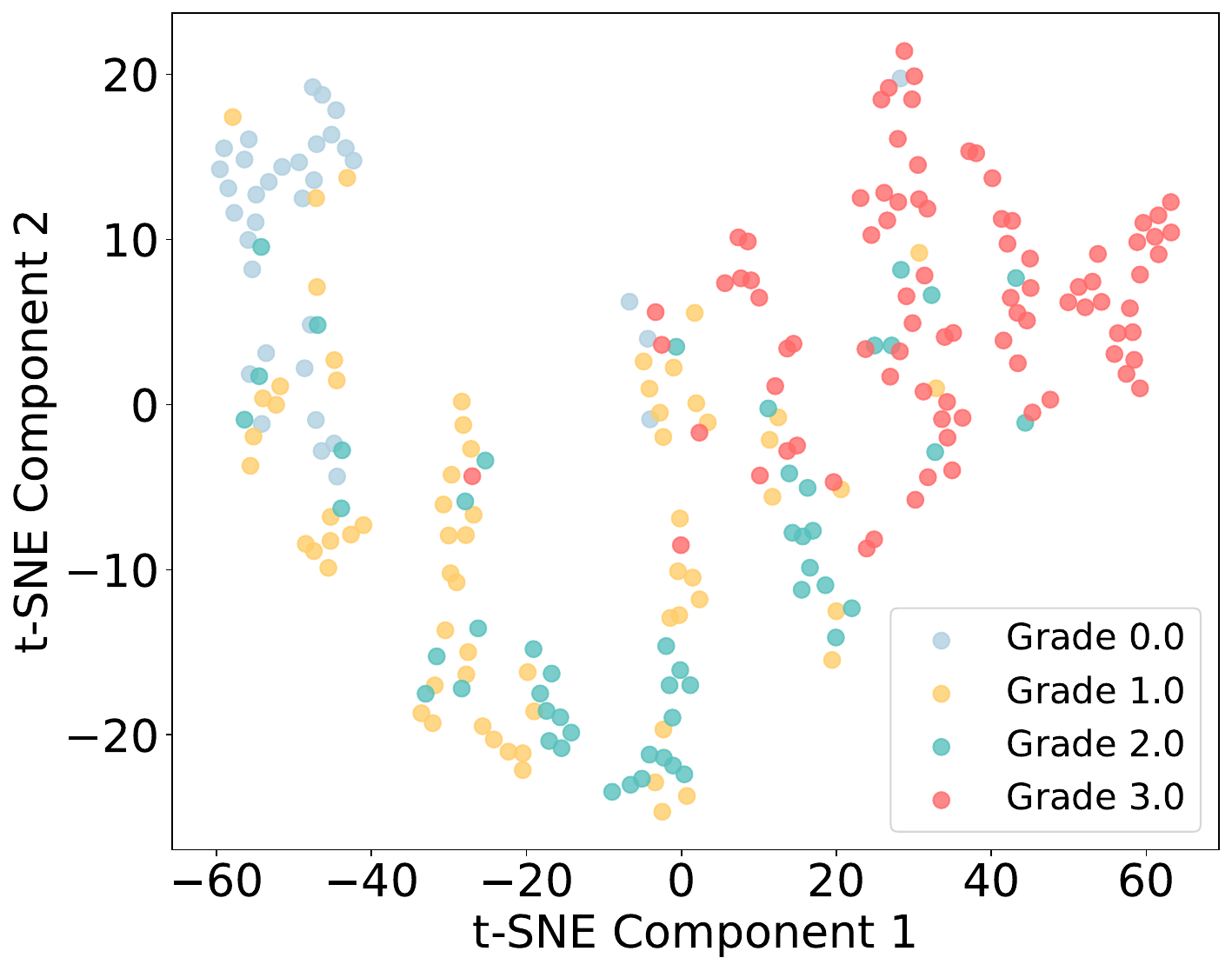}
			\label{fig:subim3}
		\end{subfigure}
		
		\vskip\baselineskip
		\vspace{-20pt}
		
		\begin{subfigure}[b]{0.3\linewidth}
			\centering
			\includegraphics[width=\linewidth]{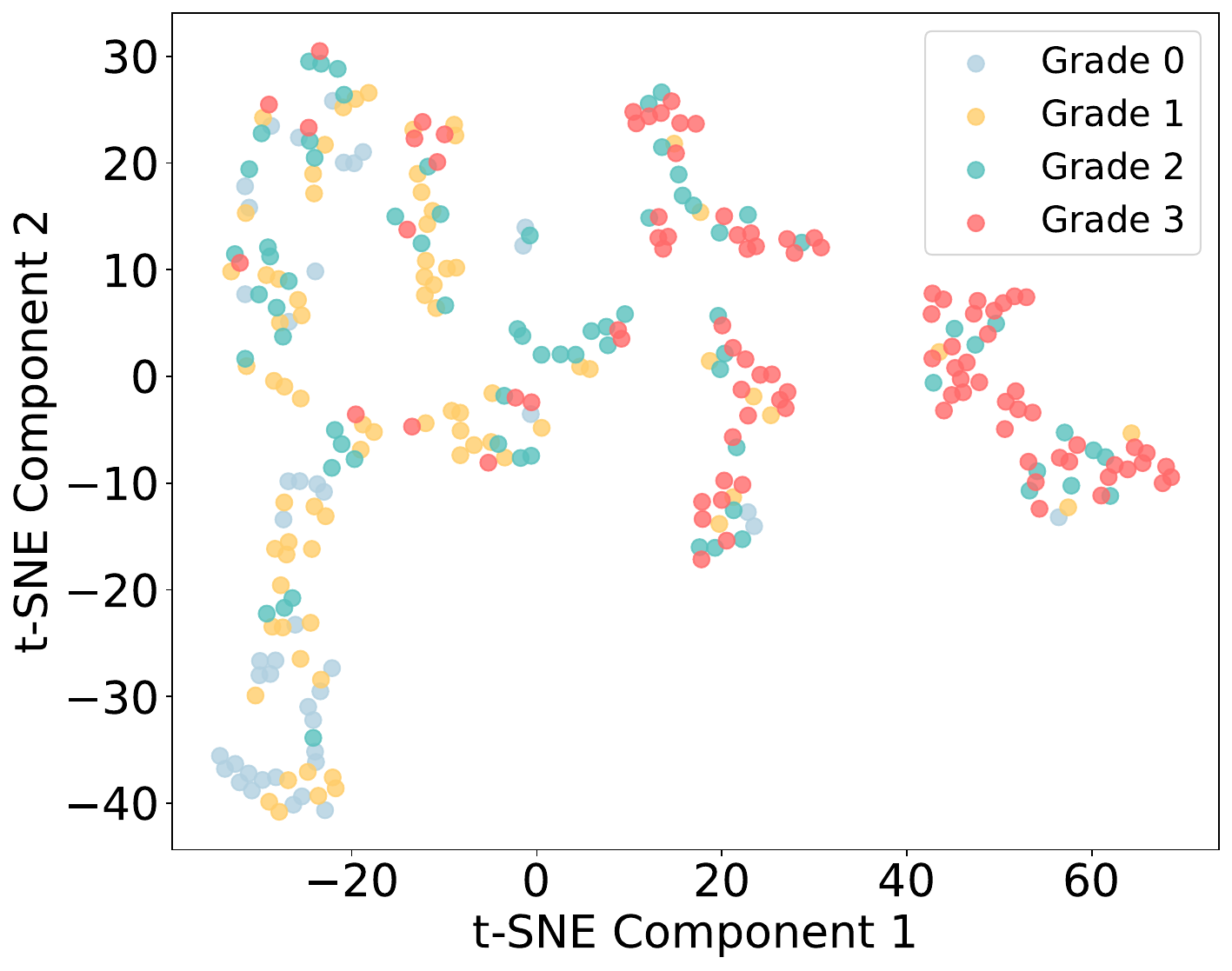}
			\caption{}
		\end{subfigure}
		\hfill
		\begin{subfigure}[b]{0.3\linewidth}
			\centering
			\includegraphics[width=\linewidth]{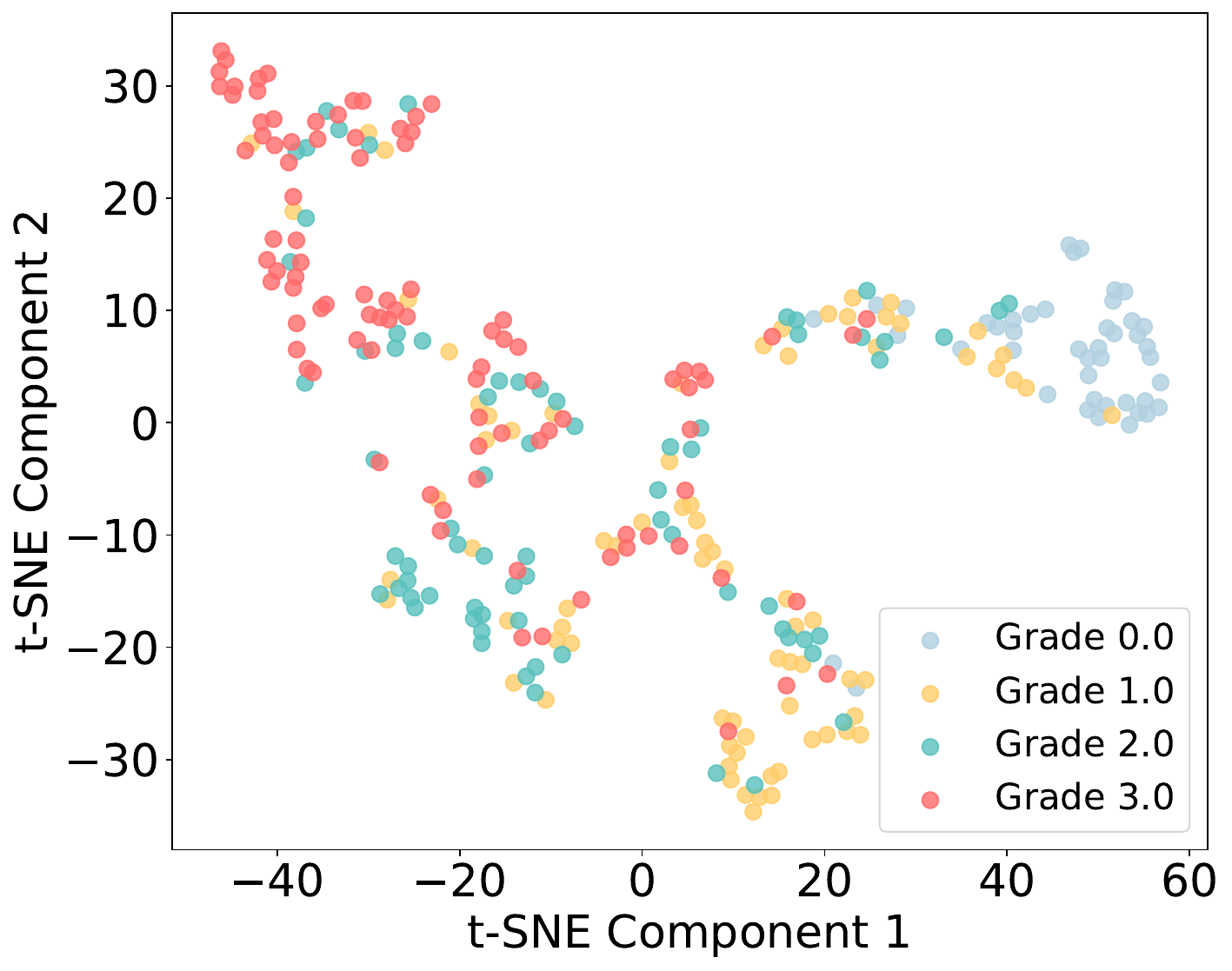}
			\caption{}
		\end{subfigure}
		\hfill
		\begin{subfigure}[b]{0.3\linewidth}
			\centering
			\includegraphics[width=\linewidth]{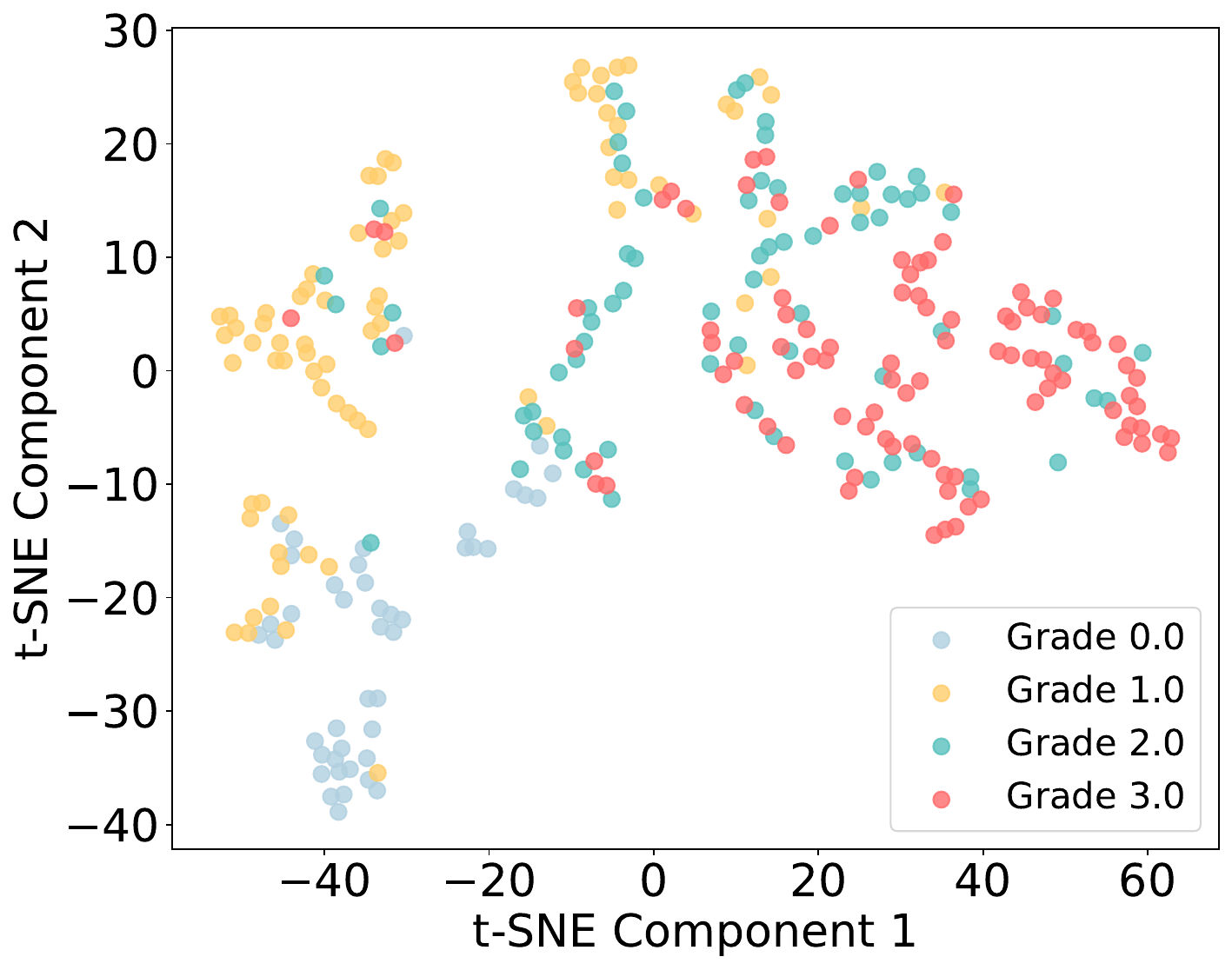}
			\caption{}
		\end{subfigure}
		
		\caption{Visualization of t-SNE applied to the validation (top) and test (bottom) datasets: (a) Single-organ esophagus, (b) Multi-organs, (c) MOON++.
		}
		\label{fig:TSNE}
	\end{figure}

	Figure~\ref{fig:TSNE} visualizes t-SNE~\citep{van2008visualizing} analysis across approaches. The single organ esophagus shows basic grade separation, with grade 3 clustering right and lower grades left. Multi-organ approach improves structure, with Grade 3 clustering top and Grade 0 bottom. The multi-modal approach (MOON++) achieves the most comprehensive separation with consistent grade progression across validation and test sets. This progressive improvement culminates in the multi-modal approach showing the most robust clustering, particularly for extreme grades (G0 and G3), while better handling intermediate grade overlaps.
	
	\begin{table*}[ht!]\caption{Comparison of different clinical prior embedding methods. ACC(\%): Accuracy Values; AUC: Area under Curve Values. Multi. ACC(\%): Multi-class Accuracy. $T_K$(\%): Kendall's tau coefficient. }
		\resizebox{\textwidth}{!}{
			\begin{tabular}{l|l|cccc|cccc}
				\hline
				\multicolumn{1}{c|}{\multirow{3}{*}{Methods}} & \multirow{3}{*}{Embeding Method} & \multicolumn{4}{c|}{Validation ($n=239$)}                                                                                                                                                      & \multicolumn{4}{c}{Independent Test ($n=289$)}                                                                                                                                          \\ \cline{3-10} 
				\multicolumn{1}{c|}{}                        &                                  & \multicolumn{1}{c|}{$\geq$G1}              & \multicolumn{1}{c|}{$\geq$G2}                         & \multicolumn{1}{c|}{G3}                               & \multirow{2}{*}{Multi. ACC/$T_K$} & \multicolumn{1}{c|}{$\geq$G1}                & \multicolumn{1}{c|}{$\geq$G2}                         & \multicolumn{1}{c|}{G3}                      & \multirow{2}{*}{Multi. ACC/$T_K$} \\ \cline{3-5} \cline{7-9}
				\multicolumn{1}{c|}{}                        &                                  & \multicolumn{1}{c|}{ACC / AUC}             & \multicolumn{1}{c|}{ACC / AUC}                        & \multicolumn{1}{c|}{ACC / AUC}                        &                                   & \multicolumn{1}{c|}{ACC / AUC}               & \multicolumn{1}{c|}{ACC / AUC}                        & \multicolumn{1}{c|}{ACC / AUC}               &                                   \\ \hline
				\multirow{6}{*}{Multi-Organs}                & -                                & \multicolumn{1}{c|}{$\textbf{93.7}/0.959$} & \multicolumn{1}{c|}{$80.3/0.856$}                     & \multicolumn{1}{c|}{$76.5/0.856$}                     & $60.3/63.5$                       & \multicolumn{1}{c|}{$95.5/0.985$}            & \multicolumn{1}{c|}{$76.8/0.862$}                     & \multicolumn{1}{c|}{$80.6/0.861$}            & $62.6/63.9$                       \\
				& CLIP(img)                        & \multicolumn{1}{c|}{$92.0 / 0.952$}        & \multicolumn{1}{c|}{$81.2 / 0.891$}                   & \multicolumn{1}{c|}{$89.1 / 0.923$}                   & $65.3 / 70.2$                     & \multicolumn{1}{c|}{$95.7 / 0.981$}          & \multicolumn{1}{c|}{$81.9 / 0.879$}                   & \multicolumn{1}{c|}{$80.0 /0.876$}           & $64.7 /71.4$                      \\
				& CLIP(text)                       & \multicolumn{1}{c|}{$91.6/0.966$}          & \multicolumn{1}{c|}{$82.0/0.865$}                     & \multicolumn{1}{c|}{$82.8/0.865$}                     & $63.6/67.9$                       & \multicolumn{1}{c|}{$94.1/0.977$}            & \multicolumn{1}{c|}{$79.2/0.850$}                     & \multicolumn{1}{c|}{$79.5/ 0.864$}           & $61.2/68.4$                       \\
				& One-hot only                     & \multicolumn{1}{c|}{$92.1/0.971$}          & \multicolumn{1}{c|}{$80.3/0.863$}                     & \multicolumn{1}{c|}{$82.8/0.907$}                     & $65.3/67.1$                       & \multicolumn{1}{c|}{$95.1/\textbf{0.986}$}   & \multicolumn{1}{c|}{$79.2/0.894$}                     & \multicolumn{1}{c|}{$\textbf{81.6}/0.895$}   & $63.1/69.8$                       \\
				& CLIP(img)+One-hot                & \multicolumn{1}{c|}{$92.4/\textbf{0.973}$} & \multicolumn{1}{c|}{$81.3/0.896$}                     & \multicolumn{1}{c|}{$81.4/0.923$}                     & $66.2/71.2$                       & \multicolumn{1}{c|}{$94.1/0.979$}            & \multicolumn{1}{c|}{$81.3/0.898$}                     & \multicolumn{1}{c|}{$80.9/0.889$}            & $64.6/71.6$                       \\
				& CLIP(img+text)                   & \multicolumn{1}{c|}{$92.8 / 0.962$}        & \multicolumn{1}{c|}{$\textbf{82.0} / \textbf{0.900}$} & \multicolumn{1}{c|}{$\textbf{91.2} / \textbf{0.959}$} & $\textbf{69.8} / \textbf{76.1}$   & \multicolumn{1}{c|}{$\textbf{96.1} / 0.986$} & \multicolumn{1}{c|}{$\textbf{84.0} / \textbf{0.921}$} & \multicolumn{1}{c|}{$80.2 / \textbf{0.894}$} & $\textbf{65.3} / \textbf{74.4}$   \\ \hline
		\end{tabular}}\label{embed}
	\end{table*}
	
	\begin{table*}[ht!]
		\caption{Comparison of different output dimension of the adaptor of CLIP image encoder. ACC(\%): Accuracy Values; AUC: Area under Curve Values. Multi. ACC(\%): Multi-class Accuracy. $T_K$(\%): Kendall's tau coefficient.}
		\resizebox{\textwidth}{!}{
			\begin{tabular}{c|c|c|cccc|cccc}
				\hline
				\multirow{3}{*}{Methods}       & \multirow{3}{*}{Embeding Methods} & \multirow{3}{*}{Adaptor Size} & \multicolumn{4}{c|}{Validation ($n=239$)}                                                                                                                                      & \multicolumn{4}{c}{Independent Test ($n=289$)}                                                                                                                                 \\ \cline{4-11} 
				&                                   &                               & \multicolumn{1}{c|}{$\geq$G1}                & \multicolumn{1}{c|}{$\geq$G2}                & \multicolumn{1}{c|}{G3}                      & \multirow{2}{*}{Multi. ACC/$T_K$} & \multicolumn{1}{c|}{$\geq$G1}                & \multicolumn{1}{c|}{$\geq$G2}                & \multicolumn{1}{c|}{G3}                      & \multirow{2}{*}{Multi. ACC/$T_K$} \\ \cline{4-6} \cline{8-10}
				&                                   &                               & \multicolumn{1}{c|}{ACC / AUC}               & \multicolumn{1}{c|}{ACC / AUC}               & \multicolumn{1}{c|}{ACC / AUC}               &                                   & \multicolumn{1}{c|}{ACC / AUC}               & \multicolumn{1}{c|}{ACC / AUC}               & \multicolumn{1}{c|}{ACC / AUC}               &                                   \\ \hline
				\multirow{5}{*}{Multi-Organs} & \multirow{5}{*}{CLIP(img)}        & 512                           & \multicolumn{1}{c|}{$91.6 / 0.945$}          & \multicolumn{1}{c|}{$79.4 / 0.852$}          & \multicolumn{1}{c|}{$81.1 / 0.856$}          & $60.6 / 64.0$                     & \multicolumn{1}{c|}{$95.5 / 0.981$}          & \multicolumn{1}{c|}{$80.6 / 0.865$}          & \multicolumn{1}{c|}{$77.1 / 0.869$}          & $59.8 / 65.3$                     \\
				&                                   & 256                           & \multicolumn{1}{c|}{$\textbf{94.5} / \textbf{0.973}$} & \multicolumn{1}{c|}{$80.7 / 0.866$}          & \multicolumn{1}{c|}{$80.3 / 0.861$}          & $64.4 / 65.3$                     & \multicolumn{1}{c|}{$95.5 / \textbf{0.987}$} & \multicolumn{1}{c|}{$78.5 / 0.870$}          & \multicolumn{1}{c|}{$80.2 / 0.868$}          & $64.4 / 65.4$                     \\
				&                                   & 128                           & \multicolumn{1}{c|}{$90.3 / 0.961$}          & \multicolumn{1}{c|}{$80.7 / 0.875$}          & \multicolumn{1}{c|}{$83.6 / 0.870$}          & $61.9 / 68.3$                     & \multicolumn{1}{c|}{$93.0 / 0.979$}          & \multicolumn{1}{c|}{$\textbf{82.0} / \textbf{0.889}$} & \multicolumn{1}{c|}{$\textbf{82.0} / 0.872$} & $64.2 / 68.4$                     \\
				&                                   & 64                            & \multicolumn{1}{c|}{$92.0 / 0.952$}          & \multicolumn{1}{c|}{$\textbf{81.2} / \textbf{0.891}$} & \multicolumn{1}{c|}{$\textbf{89.1} / \textbf{0.923}$} & $\textbf{65.3} / \textbf{70.2}$   & \multicolumn{1}{c|}{$\textbf{95.7} / 0.981$} & \multicolumn{1}{c|}{$81.9 / 0.879$}          & \multicolumn{1}{c|}{$80.0 / \textbf{0.876}$} & $\textbf{64.7} / \textbf{71.4}$   \\
				&                                   & 32                            & \multicolumn{1}{c|}{$91.8 / 0.948$} & \multicolumn{1}{c|}{$79.0 / 0.865$}          & \multicolumn{1}{c|}{$83.2 / 0.891$}          & $63.2 / 68.1$                     & \multicolumn{1}{c|}{$93.0 / 0.964$}          & \multicolumn{1}{c|}{$80.1 / 0.852$}          & \multicolumn{1}{c|}{$79.2 / 0.862$}          & $61.3 / 67.2$                     \\ \hline
		\end{tabular}}\label{clip_img}
	\end{table*}
	
	\begin{table*}[ht!]
		\caption{Comparison of different output dimension of the adaptor of CLIP image and text encoder. ACC(\%): Accuracy Values; AUC: Area under Curve Values. Multi. ACC(\%): Multi-class Accuracy. $T_K$(\%): Kendall's tau coefficient. }
		\resizebox{\textwidth}{!}{
			\begin{tabular}{c|c|c|cccc|cccc}
				\hline
				\multirow{3}{*}{Methods}       & \multirow{3}{*}{Embeding Methods} & \multirow{3}{*}{Adaptor Size} & \multicolumn{4}{c|}{Validation ($n=239$)}                                                                                                                              & \multicolumn{4}{c}{Independent Test ($n=289$)}                                                                                                                        \\ \cline{4-11} 
				&                             &                               & \multicolumn{1}{c|}{$\geq$G1}        & \multicolumn{1}{c|}{$\geq$G2}        & \multicolumn{1}{c|}{G3}               & \multirow{2}{*}{Multi. ACC/$T_K$} & \multicolumn{1}{c|}{$\geq$G1}        & \multicolumn{1}{c|}{$\geq$G2}        & \multicolumn{1}{c|}{G3}               & \multirow{2}{*}{Multi. ACC/$T_K$} \\ \cline{4-6} \cline{8-10} 
				&                             &                               & \multicolumn{1}{c|}{ACC / AUC}       & \multicolumn{1}{c|}{ACC / AUC}       & \multicolumn{1}{c|}{ACC / AUC}        &                                   & \multicolumn{1}{c|}{ACC / AUC}       & \multicolumn{1}{c|}{ACC / AUC}       & \multicolumn{1}{c|}{ACC / AUC}        &                                   \\ \hline
				\multirow{3}{*}{Multi-Organs} & \multirow{3}{*}{CLIP(img+text)} & 128+128                       & \multicolumn{1}{c|}{$92.5/0.962$}          & \multicolumn{1}{c|}{$78.7/0.880$}          & \multicolumn{1}{c|}{$89.1/0.955$}          & $66.1/71.7$                       & \multicolumn{1}{c|}{$95.5/0.975$}          & \multicolumn{1}{c|}{$82.4/0.919$}          & \multicolumn{1}{c|}{\textbf{80.6}/0.892}       & $64.6/\textbf{74.5}$                \\
				&                             & 64+64                         & \multicolumn{1}{c|}{$92.8/0.962$}          & \multicolumn{1}{c|}{\textbf{82.0/0.900}}   & \multicolumn{1}{c|}{\textbf{91.2/0.959}}   & \textbf{69.8}/\textbf{76.1}         & \multicolumn{1}{c|}{\textbf{96.1/0.986}}   & \multicolumn{1}{c|}{\textbf{84.0/0.921}}   & \multicolumn{1}{c|}{$80.2/\textbf{0.894}$}    & \textbf{65.3}/74.4                \\
				&                             & 32+32                         & \multicolumn{1}{c|}{\textbf{94.6/0.974}}   & \multicolumn{1}{c|}{$79.9/0.882$}          & \multicolumn{1}{c|}{$84.1/0.907$}          & $65.6/69.5$                       & \multicolumn{1}{c|}{$93.8/0.978$}          & \multicolumn{1}{c|}{$81.3/0.875$}          & \multicolumn{1}{c|}{\textbf{80.6}/0.884}       & $62.9/68.2$                       \\ \hline
		\end{tabular}}\label{clip_text}
	\end{table*}

	\begin{table*}[ht!]\caption{Ablation study of using different text descriptions, (N) denotes using directly the numerical values, (C) denotes using the categorical  text descriptions as shown in Table~\ref{level}. ACC(\%): Accuracy Values; AUC: Area under Curve Values. Multi. ACC(\%): Multi-class Accuracy. $T_K$(\%): Kendall's tau coefficient.}
		\resizebox{\textwidth}{!}{\begin{tabular}{c|c|l|cccc|cccc}
				\hline
				\multirow{3}{*}{Methods}       & \multirow{3}{*}{Embeding Methods} & \multicolumn{1}{c|}{\multirow{3}{*}{Text}} & \multicolumn{4}{c|}{Validation ($n=239$)}                                                                                                                                 & \multicolumn{4}{c}{Independent Test ($n=289$)}                                                                                                                           \\ \cline{4-11} 
				&                                   & \multicolumn{1}{c|}{}                      & \multicolumn{1}{c|}{$\geq$G1}               & \multicolumn{1}{c|}{$\geq$G2}              & \multicolumn{1}{c|}{G3}                    & \multirow{2}{*}{Multi. ACC/$T_K$} & \multicolumn{1}{c|}{$\geq$G1}              & \multicolumn{1}{c|}{$\geq$G2}              & \multicolumn{1}{c|}{G3}                    & \multirow{2}{*}{Multi. ACC/$T_K$} \\ \cline{4-6} \cline{8-10}
				&                                   & \multicolumn{1}{c|}{}                      & \multicolumn{1}{c|}{ACC / AUC}              & \multicolumn{1}{c|}{ACC / AUC}             & \multicolumn{1}{c|}{ACC / AUC}             &                                   & \multicolumn{1}{c|}{ACC / AUC}             & \multicolumn{1}{c|}{ACC / AUC}             & \multicolumn{1}{c|}{ACC / AUC}             &                                   \\ \hline
				\multirow{4}{*}{Multi-Organs} & \multirow{4}{*}{CLIP(img+text)}   & Volume (N)                                 & \multicolumn{1}{c|}{$92.9/0.963$}           & \multicolumn{1}{c|}{$78.7 /0.866$}         & \multicolumn{1}{c|}{$87.4/0.940$}          & $66.1/70.1$                       & \multicolumn{1}{c|}{$93.1 /0.984$}         & \multicolumn{1}{c|}{$79.6/0.887$}          & \multicolumn{1}{c|}{$79.6/\textbf{0.895}$} & $62.1/67.8$                       \\
				&                                   & Volume (C)                                 & \multicolumn{1}{c|}{$93.3/\textbf{0.975}$}  & \multicolumn{1}{c|}{$\textbf{82.8}/0.887$} & \multicolumn{1}{c|}{$90.0/0.945$}          & $\textbf{72.8}/74.6$              & \multicolumn{1}{c|}{$92.7/0.974$}          & \multicolumn{1}{c|}{$80.6/0.878$}          & \multicolumn{1}{c|}{$78.5/0.864$}          & $62.5/68.2$                       \\
				&                                   & Volume (C)+ LSVR (N)                       & \multicolumn{1}{c|}{$\textbf{94.6}/0.962 $} & \multicolumn{1}{c|}{$80.3/0.867$}          & \multicolumn{1}{c|}{$87.4/0.924$}          & $69.0/70.6$                       & \multicolumn{1}{c|}{$94.5/0.978$}          & \multicolumn{1}{c|}{$82.0/0.904$}          & \multicolumn{1}{c|}{$\textbf{81.3}/0.892$} & $63.5/70.7$                       \\
				&                                   & Volume (C)+ LSVR (C)                       & \multicolumn{1}{c|}{$92.8/0.962$}           & \multicolumn{1}{c|}{$82.0/\textbf{0.900}$} & \multicolumn{1}{c|}{$\textbf{91.2/0.959}$} & $69.8/\textbf{76.1}$              & \multicolumn{1}{c|}{$\textbf{96.1/0.986}$} & \multicolumn{1}{c|}{$\textbf{84.0/0.921}$} & \multicolumn{1}{c|}{$80.2/0.894$}          & $\textbf{65.3/74.4}$              \\ \hline
			\end{tabular}
		}\label{text_des}
	\end{table*}
	
\begin{figure}[ht!]
		\centering
		\centering
		\includegraphics[width=1\linewidth]{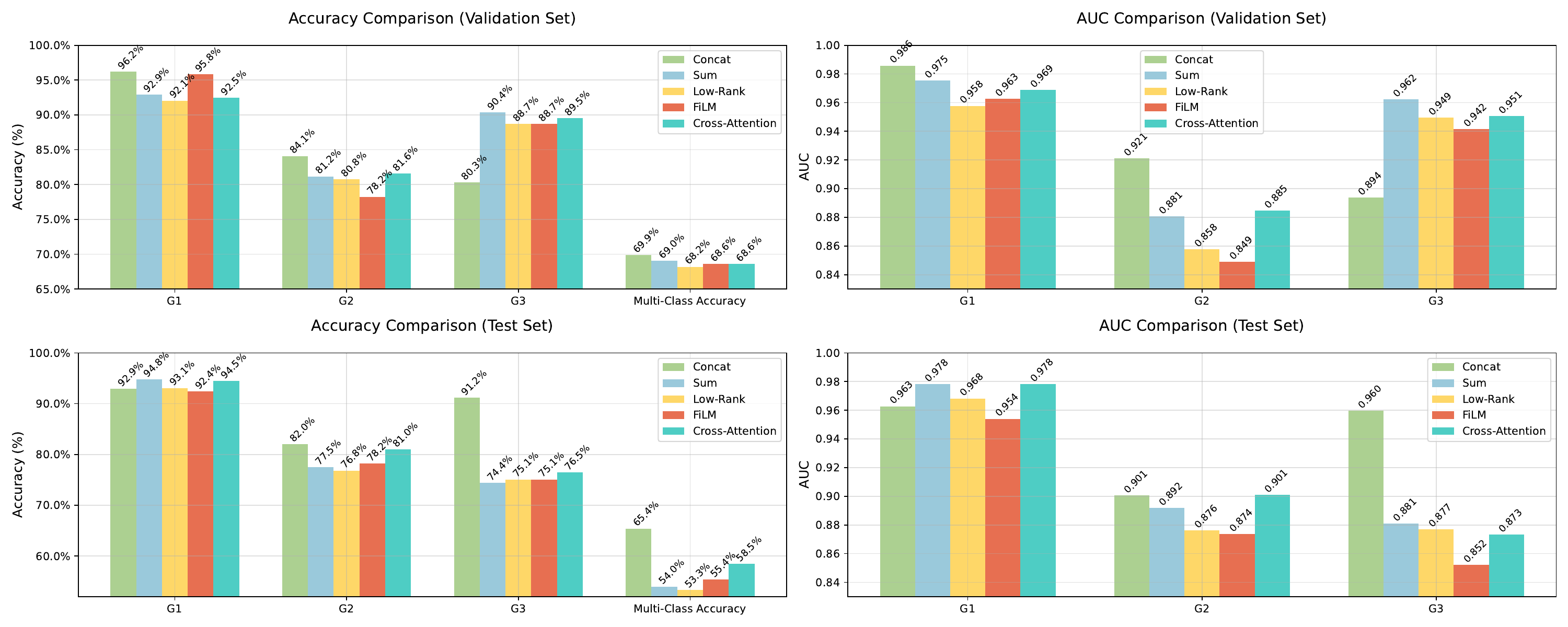}
		\caption{Comparison of different fusion strategies.}
		\label{fig:fusion}
	\end{figure}

	\begin{table*}[ht!]
		\caption{Ablation study on the impact of applying supervision to CLIP features for alignment using contrastive loss~\citep{he2020momentum} and InfoNCE loss~\citep{oord2018representation}. ACC(\%): Accuracy Values; AUC: Area under Curve Values. Multi. ACC(\%): Multi-class Accuracy. $T_K$(\%): Kendall's tau coefficient.}
		\resizebox{\textwidth}{!}{
			\begin{tabular}{c|c|l|cccc|cccc}
				\hline
				\multirow{3}{*}{Methods}       & \multirow{3}{*}{Embeding Methods} & \multicolumn{1}{c|}{\multirow{3}{*}{Supervision on $H_{img}, H_{text}$}} & \multicolumn{4}{c|}{Validation ($n=239$)}                                                                                                                              & \multicolumn{4}{c}{Independent Test ($n=289$)}                                                                                                                        \\ \cline{4-11} 
				&                             & \multicolumn{1}{c|}{}                                                    & \multicolumn{1}{c|}{$\geq$G1}        & \multicolumn{1}{c|}{$\geq$G2}        & \multicolumn{1}{c|}{G3}               & \multirow{2}{*}{Multi. ACC/$T_K$} & \multicolumn{1}{c|}{$\geq$G1}        & \multicolumn{1}{c|}{$\geq$G2}        & \multicolumn{1}{c|}{G3}               & \multirow{2}{*}{Multi. ACC/$T_K$} \\ \cline{4-6} \cline{8-10} 
				&                             & \multicolumn{1}{c|}{}                                                    & \multicolumn{1}{c|}{ACC / AUC}       & \multicolumn{1}{c|}{ACC / AUC}       & \multicolumn{1}{c|}{ACC / AUC}        &                                   & \multicolumn{1}{c|}{ACC / AUC}       & \multicolumn{1}{c|}{ACC / AUC}       & \multicolumn{1}{c|}{ACC / AUC}        &                                   \\ \hline
				\multirow{3}{*}{Multi-Organs} & \multirow{3}{*}{CLIP(img+text)} & -                                                                        & \multicolumn{1}{c|}{$92.8/0.962$}          & \multicolumn{1}{c|}{$82.0/0.900$}          & \multicolumn{1}{c|}{$91.2/0.959$}          & \textbf{69.8}/\textbf{76.1}         & \multicolumn{1}{c|}{\textbf{96.1/0.986}}   & \multicolumn{1}{c|}{$84.0/\textbf{0.921}$}    & \multicolumn{1}{c|}{$80.2/\textbf{0.894}$}    & \textbf{65.3/74.4}                \\
				&                             & Contrastive                                       & \multicolumn{1}{c|}{$93.5/\textbf{0.968}$}    & \multicolumn{1}{c|}{\textbf{82.5/0.902}}   & \multicolumn{1}{c|}{$90.2/0.946$}          & $67.5/74.2$                       & \multicolumn{1}{c|}{$94.3/0.965$}          & \multicolumn{1}{c|}{$83.6/0.918$}          & \multicolumn{1}{c|}{$83.2/0.886$}          & $64.5/70.3$                       \\
				&                             & Contrastive +InfoNCE & \multicolumn{1}{c|}{\textbf{93.6}/0.935}       & \multicolumn{1}{c|}{$82.3/0.896$}          & \multicolumn{1}{c|}{\textbf{91.7/0.961}}   & $68.8/74.9$                       & \multicolumn{1}{c|}{$95.5/0.972$}          & \multicolumn{1}{c|}{\textbf{84.6}/0.920}       & \multicolumn{1}{c|}{\textbf{83.4}/0.884}       & $64.7/73.8$                       \\ \hline
		\end{tabular}}\label{supervision}
	\end{table*}
	\noindent\textbf{Post Fusion.} 
	We conduct ablation study on different fusion strategies as shown in Figure~\ref{fig:fusion}, comparing concatenation~\citep{feichtenhofer2016convolutional}, pred-sum~\citep{simonyan2014two}, low-rank fusion~\citep{liu2018efficient}, FiLM~\citep{perez2018film}, and cross-attention~\citep{vaswani2017attention}. 
	On the validation set, all strategies demonstrate strong G1 classification performance (92-96\% accuracy), with concatenation achieving the highest accuracy (96\%) and AUC (0.98). G2 classification shows a wider performance variation (75-84\%), while the G3 classification exhibits consistent performance across strategies (88-90\%). The results of the test set maintain high accuracy in classification of G1 (92-94\%), but show a degradation in performance for the classification of G2. In particular, concatenation maintains superior performance for G3 classification ($\sim$91\%) while other strategies cluster around 75\%.
	
	The concatenation strategy demonstrates the most robust performance across both validation (68-70\% multi-class accuracy) and test sets (53-65\% multi-class accuracy), showing better generalization compared to other approaches. These results indicate that the straightforward concatenation strategy provides the most reliable fusion approach to maintain consistent performance across different datasets and severity grades. Therefore, we adopted concatenation for fusing multi-modal features.
	\subsection{Reader Study}
	
	\begin{figure}[ht!]
		\centering
		\begin{subfigure}[b]{\linewidth}
			\centering
			\includegraphics[width=\linewidth]{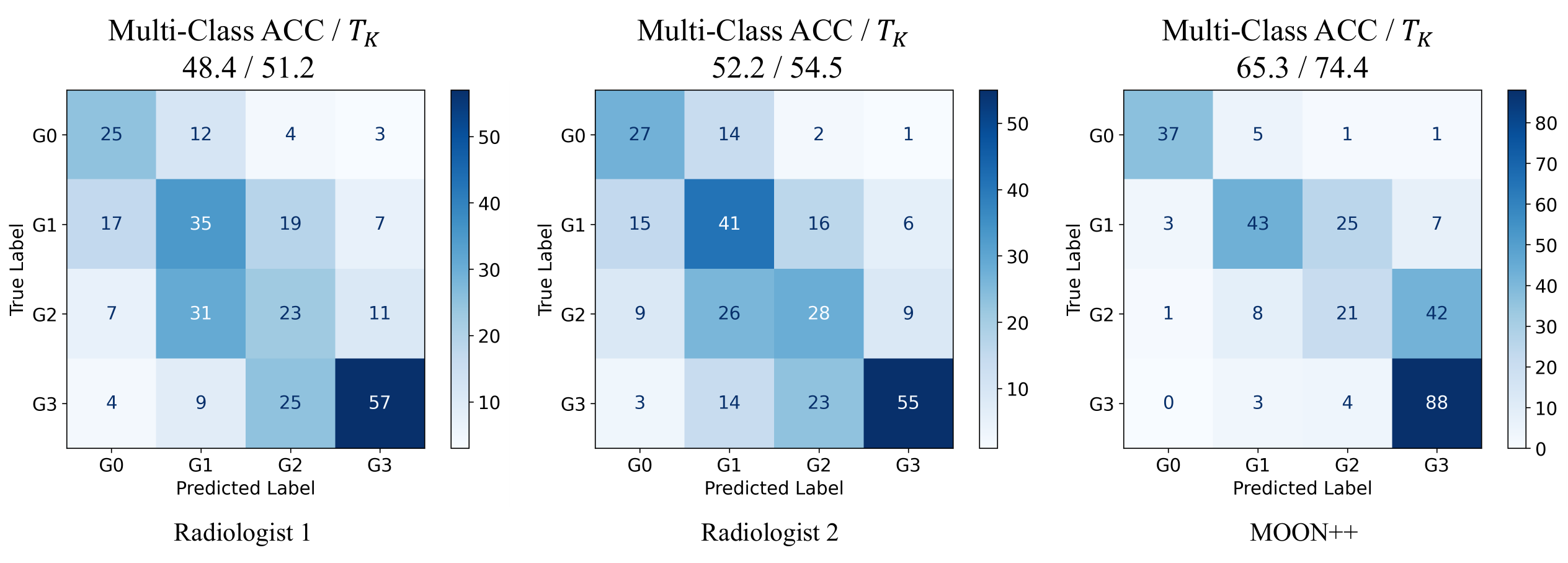}
			\label{subfig:confu_matrix}
		\end{subfigure}
		\vspace{10pt}
		\begin{subfigure}[b]{\linewidth}
			\centering
			\includegraphics[width=\linewidth]{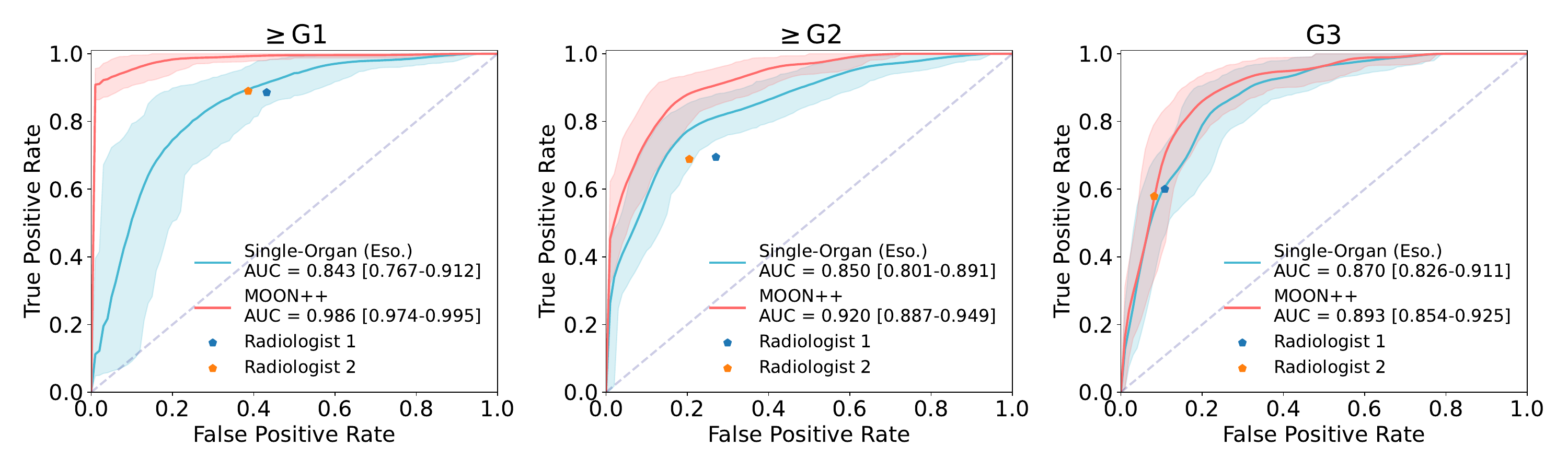}
			
			\label{subfig:roc_curve}
		\end{subfigure}
		
		\begin{subfigure}{\linewidth}
			\centering
			\resizebox{\textwidth}{!}{%
				\begin{tabular}{l c c c c}
					\toprule
					Class & Precision & Recall & F1-Score & Support \\
					\cmidrule(r){2-4}
					& \multicolumn{3}{c}{Radiologist 1 / Radiologist 2 / MOON++} & (N) \\
					\midrule
					G0 & 0.472 / 0.500 / \textbf{0.902} & 0.568 / 0.614 / \textbf{0.841} & 0.515 / 0.551 / \textbf{0.871} & 44 \\
					G1 & 0.402 / 0.432 / \textbf{0.729} & 0.449 / 0.526 / \textbf{0.551} & 0.424 / 0.474 / \textbf{0.628} & 78 \\
					G2 & 0.324 / 0.406 / \textbf{0.412} & 0.319 / \textbf{0.389} / 0.292 & 0.322 / \textbf{0.397} / 0.341 & 72 \\
					G3 & 0.731 / \textbf{0.775} / 0.638 & 0.600 / 0.579 / \textbf{0.926} & 0.659 / 0.663 / \textbf{0.755} & 95 \\
					\bottomrule
				\end{tabular}
			}
		\end{subfigure}
		\caption{Comprehensive reader study results. (top) Confusion matrices visualizing classification results, (middle) Receiver operating characteristic (ROC) curves comparing diagnostic performance, 95\% confidence intervals were calculated using 1,000 bootstrap replications, (bottom) Detailed performance metrics, with the best value in each category highlighted in bold.}
		\label{fig:reader}
		
	\end{figure}

	\begin{figure*}[ht!]
		\centering
		\centering
		\includegraphics[width=1\linewidth]{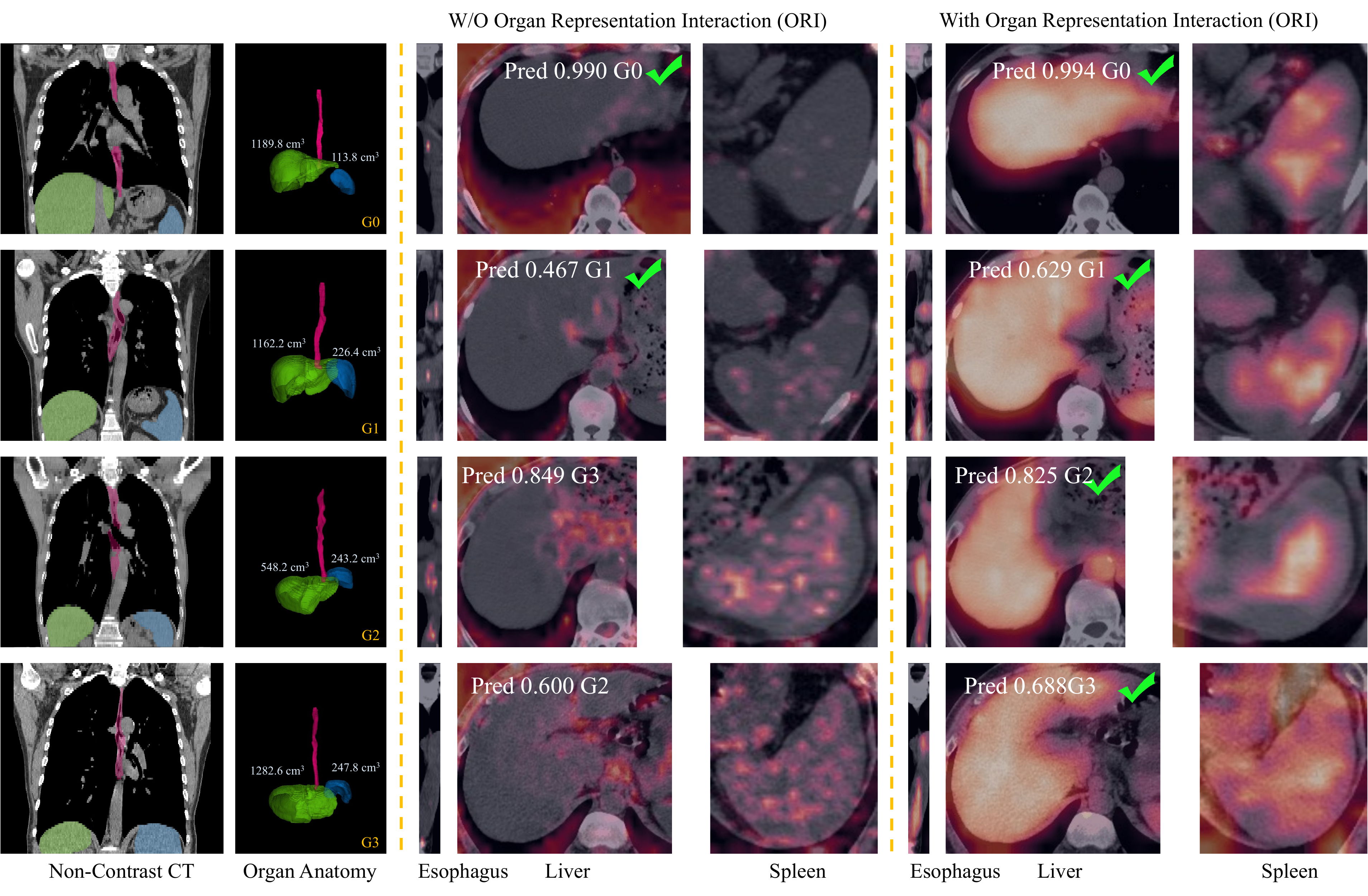}
		\caption{Comparison of Grad-CAM visualizations across different organs.}
		\label{fig:CAM}
	\end{figure*}

	\begin{figure}[ht!]
		\centering
		\centering
		\includegraphics[width=1\linewidth]{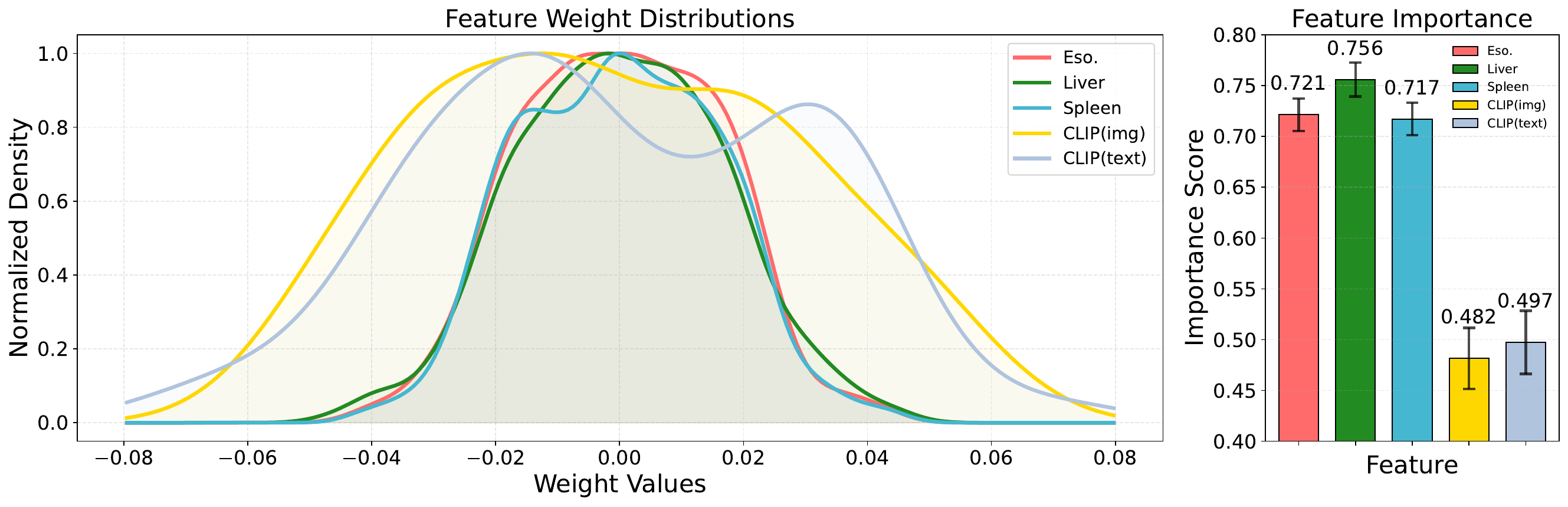}
		\caption{Visualization of feature importance based on the final classifier's learned weights. (Left) Normalized density distributions of the weights corresponding to each feature group, smoothed using Gaussian Kernel Density Estimation. (Right) Corresponding importance scores calculated as the L2-norm of the weight submatrices, where a larger norm indicates greater feature influence.}
		\label{fig:GKD}
	\end{figure} 
	
	\begin{figure}[ht!]
		\centering
		\centering
		\includegraphics[width=1\linewidth]{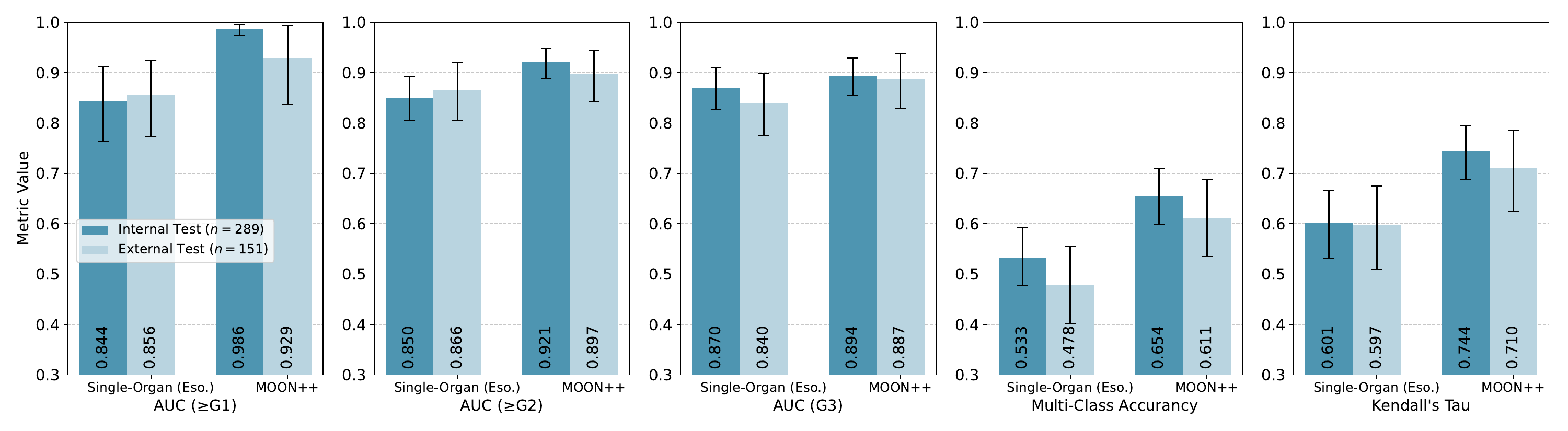}
		\caption{Performance comparison between internal test set ($n=289$) and external validation set ($n=151$). Error bars indicate 95\% confidence intervals. AUC: Area Under Curve Values.}
		\label{fig:ext}
	\end{figure}

	A comprehensive reader study was conducted with two experienced radiologists, as illustrated in Figure~\ref{fig:reader}, to assess the clinical efficacy of MOON++. Radiologist 1 and Radiologist 2 had $>$5 and $>$10 years of clinical experience, respectively. Radiologists followed the staging definition of EV as described in the guidelines of the ESGE~\citep{gralnek2022endoscopic} and JSPH~\citep{kawano2008short} guidelines, focusing on esophageal morphology and drawing from their experience with surrounding organs. The model achieved a multi-class accuracy of 65.3\% and a Kendall's tau of 74.4\%, surpassing the performance of the radiologists, who recorded scores of 48.4\%/51.2\% and 52.2\%/54.5\% respectively. In the confusion matrix analysis, the model was particularly effective in classifying G3 cases (severe), accurately predicting 88 instances compared to the radiologists' predictions of 57 and 55, while maintaining low false negatives with only seven misclassified cases. Furthermore, the model demonstrated improved differentiation between the G0 and G1 grades, accurately identifying 37 G0 and 43 G1 cases, while radiologists exhibited greater confusion in these categories, as indicated by increased off-diagonal elements in their matrices. 
	
	The analysis of the ROC curve further supports the clinical utility of our model across various classification severity thresholds. For G1 detection, the model achieved an AUC of 0.986, outperforming the single-organ approach's 0.844, with both radiologists' performance points below the multi-modal ROC curve. This advantageous performance continued for G2 detection, where the model achieved an AUC of 0.921, illustrating a notable advantageous performance gap compared to radiologists. In G3 detection, the model sustained strong performance with a multi-modal AUC of 0.894, aligning more closely with radiologist predictions for these critical cases. MOON++ significantly outperformed both radiologists in classifying $\geq$ G1 (96.2\% vs. 83.7\%, 84.8\%; $p < 0.001$) and $\geq$ G2 (84.1\% vs. 70.9\%, 73.4\%; $p < 0.001$, $p < 0.01$). For G3 classification, all performances were comparable (80.3\% vs. 79.6\%, 80.6\%; $p > 0.05$), with no significant differences between radiologists across all grades.

	In general, these findings highlight the capabilities of the model in addressing clinical challenges by reducing inter-observer variability, maintaining consistent grading across severity levels, providing improved discrimination between adjacent grades, and ensuring high sensitivity for severe cases while minimizing false positives. These results indicate the model's good potential to enhance clinical decision-making and EV grading standardization in practical real-world applications.
	
	\subsection{Visualization}
	\noindent\textbf{Feature Interaction.} 
	Figure~\ref{fig:CAM} presents Grad-CAM~\citep{selvaraju2017grad} visualizations comparing non-interactive and interactive feature representations across EV severity grades (G0-G3) in test set. While non-interactive features show scattered, localized activation patterns, interactive features demonstrate broader, more coherent attention regions with smoother transitions. As severity progresses from G0 to G3, interactive features exhibit clearer grade differentiation and more extensive activation patterns, particularly in higher grades. Organ-specific analysis reveals that interactive features improve anatomical coverage in the liver, enhance spleen alignment, and offer better delineation of esophageal tissue involvement. Interactive features enhance inter-organ correlation and reduce noise, improving the detection of pathologically significant regions and overall disease severity assessment.

	\noindent\textbf{Feature Importance.} 
	Figure~\ref{fig:GKD} presents a detailed visualization of feature weight distributions and importance scores in the final classifier. The left plot reveals distinct patterns between organ and CLIP features. Image-based organ features demonstrate narrow, concentrated Gaussian distributions, with peak normalized densities of 0.95-1.0 within a symmetric range. In contrast, CLIP-like 3D image and text features exhibit broader distributions with higher variance, starting from -0.08 and showing right-skewed tendencies, with lower peak normalized densities. The right plot quantifies the importance of features in all modalities, where image-based organ features demonstrate higher importance scores. These organ characteristics consistently maintain scores above 0.7 with minimal variance, as indicated by their small error bars. CLIP modalities show markedly lower importance, with text and image features scoring 0.497 and 0.482 respectively suggesting more variable contributions. This lower importance compared to organ features indicates their supplementary rather than primary role.
	
	These findings establish a clear hierarchy in feature importance, emphasizing that while multi-modal fusion offers benefits, organ-specific features remain the primary determinants of model performance. Image-based features demonstrate stronger predictive weight compared to general vision language features, suggesting that the multi-organ analysis forms the foundation of accurate EV assessment.

	\subsection{External Validation}
	To further validate the generalization of proposed methods, we collected an external validation set  using an external validation dataset comprising 151 patients from the Second Affiliated Hospital of Baotou Medical College, Inner Mongolia, China. The comparative results between internal test ($n=289$) and external validation ($n=151$) are presented in Figure~\ref{fig:ext}. Statistical analysis using permutation tests revealed no significant differences ($p>0.05$) between the internal and external test sets across performance metrics.
	\section{Discussion \& Conclusion}
	
	In this work, we present MOON++, a novel multi-modal approach for EV classification using non-contrast CT scans, achieving significant performance improvements over single-organ methods, with 65.3\% multi-class accuracy and a Kendall's tau of 74.4\%, exceeding both radiologist quantitative performance and previous approaches. MOON++ excels in multi-class accuracy when compared to radiologists (65.3\% vs. 48.4\%/52.2\%), especially in effectively identifying clinically crucial G3 patient cases.  Following the updated Baveno VII consensus guidelines~\citep{de2022baveno}, accurate EV staging enables appropriate treatment selection, from vasoactive drugs and non-selective beta-blockers to endoscopic band ligation, aimed at preventing bleeding events and further hepatic decompensation. The use of NCCT as a less invasive alternative to endoscopy and CE-CT may facilitate patient monitoring throughout the treatment process, reducing the need for repeated invasive procedures.
	
	The high precision in G0 classification is attributed to the ordinal regression approach and the inclusion of certain normal subjects without liver fibrosis, showing F1-score of 0.871, highlighting the need for further validation with subjects who have liver fibrosis but do not have EV to ensure clinical robustness. Feature importance analysis identifies organ-specific characteristics as primary predictors (liver, esophagus, spleen), with CLIP features providing complementary information, reflecting clinical insights into disease progression. 
	
	MOON++ emphasizes the synergy of domain knowledge and advances deep learning techniques to tackle complex clinical challenges, underscoring the importance of balanced feature representations that uphold the clinical relevance and cross-modal alignment for effective EV grading. Our investigation of clinical information embedding demonstrates that while simple one-hot encoding of organ volumes and LSVR achieves substantial improvements (63.1\% accuracy), medical MLLM provides smoother embedding and better CT-text alignment, leading to further enhanced performance (65.3\% accuracy).
	
	Challenges remain in distinguishing intermediate grades, with relatively low F1-scores (G1: 0.628, and G2: 0.341). The multi-modal performance relies on segmentation quality for computing clinical priors. Future research should focus on enhancing discrimination between these grades through large-scale multi-center validation datasets, including more cases of liver fibrosis without EV, and explore applications to other multi-organ conditions. MOON++ emphasizes the synergy of domain knowledge and advances deep learning techniques to tackle complex clinical challenges, underscoring the importance of balanced feature representations that uphold the clinical relevance and cross-modal alignment for effective EV grading. 
	
	In summary, we present MOON++, a multi-modal framework for noninvasive evaluation of esophageal varices using non-contrast CT scans. Inspired by radiologists' diagnostic processes, this approach integrates features from multiple organs involved in EV development, incorporating clinically relevant volumetric relationships through multi-modal learning. Our extensive testing on a dataset of 1,474 patients shows that MOON++ not only closes the diagnostic gap in the evaluation of EV, but also exceeds previous CE-CT-based methods, achieving superior metrics in both validation and independent test cohorts. MOON++ demonstrates potential for broader clinical use, providing a less invasive and more accessible alternative to traditional methods. Future efforts will focus on improving the model by adding clinical indicators such as blood test markers to improve classification accuracy and creating a more comprehensive EV assessment system. Furthermore, research will explore the applicability of the model in various patient populations and clinical settings to ensure its greater generalizability.
	\section*{Conflict of Interest}
	The authors have no competing interests to declare that are relevant to the content of this article.
	\section*{CRediT authorship contribution statement}
	\textbf{X. Zhang:} Software, Methodology, Formal analysis, Writing – original draft. \textbf{C. Li:} Conceptualization, Data curation, Methodology, Writing – original draft. \textbf{J. Hao:} Methodology, Software. \textbf{Y. Gao:} Methodology, Formal analysis. Writing - review \& editing. \textbf{D. Tu:} Methodology, Formal analysis. \textbf{J. Qiao:} Conceptualization, Data curation. \textbf{X. Yin:} Conceptualization, Data curation. \textbf{L. Lu:} Supervision, Writing - review \& editing. \textbf{L. Zhang:} Supervision, Writing - review \& editing. \textbf{K. Yan:} Supervision, Writing - review \& editing. \textbf{Y. Hou:} Conceptualization, Funding acquisition, Data curation \textbf{Y. Shi:} Conceptualization, Funding acquisition, Data curation, Writing – review \& editing.
	\section*{Acknowledgments}
	The authors thank Professor Maria A. Zuluaga of EURECOM and King's College London for her insightful suggestions and valuable discussions for the manuscript's revision. The authors also thank the Second Affiliated Hospital of Baotou Medical College and Dr. Qi Liu of the Department of Radiology for their contributions to the external clinical validation.
	
	This work was supported by Liaoning Province Science and Technology Joint Plan (2023JH2/101700127), the Leading Young Talent Program of Xingliao Yingcai in Liaoning Province (XLYC2203037), Liaoning Provincial Science and Technology Program (2025-BS-0581).
	%%Harvard
	
	%\bibliographystyle{model2-names.bst}\biboptions{authoryear}
	%\bibliography{strings,refs}
	\biboptions{authoryear}

\end{document}